\documentclass[sigconf]{acmart}

\author{Ronghui Xu}
\affiliation{%
    \orcid{0009-0006-9282-4171}
    \institution{Central South University}
    \city{Changsha}
    \country{China}}
\email{ronghuixu@csu.edu.cn}

\author{Hao Miao}
\affiliation{
    \orcid{0000-0001-9346-7133}
    \institution{Aalborg University}
    \city{Aalborg}
  \country{Denmark}
}
\email{haom@cs.aau.dk}

\author{Senzhang Wang}
\authornote{Corresponding author.}
\affiliation{
    \orcid{0000-0002-3615-4859}
    \institution{Central South University}
    \city{Changsha}
    \country{China}
}
\email{szwang@csu.edu.cn}

\author{Philip S. Yu}
\affiliation{
    \orcid{0000-0002-3491-5968}
    \institution{University of Illinois at Chicago}
    \city{Chicago}
    \country{USA}
}
\email{psyu@uic.edu}

\author{Jianxin Wang}
\affiliation{
    \orcid{0000-0003-1516-0480}
    \institution{Central South University}
    \city{Changsha}
    \country{China}
}
\email{jxwang@csu.edu.cn}

\usepackage[linesnumbered,ruled,vlined]{algorithm2e}
\usepackage{amsmath}
\usepackage{xcolor}
\usepackage{subcaption}
\usepackage{geometry}  
\usepackage{booktabs}  
\usepackage{siunitx}   
\usepackage{multirow}
\usepackage{float}
\usepackage{color}
\usepackage{graphicx}
\usepackage{tabularx}
\usepackage{xcolor}
\usepackage{caption}
\usepackage{makecell}
\usepackage{hyperref}
\usepackage{enumitem}
\usepackage{bm} 

\AtBeginDocument{%
  \providecommand\BibTeX{{%
    \normalfont B\kern-0.5em{\scshape i\kern-0.25em b}\kern-0.8em\TeX}}}

\copyrightyear{2024}
\acmYear{2024}
\setcopyright{acmlicensed}\acmConference[KDD '24]{Proceedings of the 30th ACM SIGKDD Conference on Knowledge Discovery and Data Mining}{August 25--29, 2024}{Barcelona, Spain}
\acmBooktitle{Proceedings of the 30th ACM SIGKDD Conference on Knowledge Discovery and Data Mining (KDD '24), August 25--29, 2024, Barcelona, Spain}

\graphicspath{{image/}}
\begin{document}


\title{PeFAD: A Parameter-Efficient Federated Framework for Time Series Anomaly Detection}


\begin{abstract}
With the proliferation of mobile sensing techniques, huge amounts of time series data are generated and accumulated in various domains, fueling plenty of real-world applications. In this setting, time series anomaly detection is practically important. It endeavors to identify deviant samples from the normal sample distribution in time series. Existing approaches generally assume that all the time series is available at a central location. However, we are witnessing the decentralized collection of time series due to the deployment of various edge devices. To bridge the gap between the decentralized time series data and the centralized anomaly detection algorithms, we propose a \underline{P}arameter-\underline{e}fficient \underline{F}ederated \underline{A}nomaly \underline{D}etection framework named PeFAD with the increasing privacy concerns.
PeFAD for the first time employs the pre-trained language model (PLM) as the body of the client's local model, which can benefit from its cross-modality knowledge transfer capability. To reduce the communication overhead and local model adaptation cost, we propose a parameter-efficient federated training module such that clients only need to fine-tune small-scale parameters and transmit them to the server for update. PeFAD utilizes a novel anomaly-driven mask selection strategy to mitigate the impact of neglected anomalies during training. A knowledge distillation operation on a synthetic privacy-preserving dataset that is shared by all the clients is also proposed to address the data heterogeneity issue across clients. 
We conduct extensive evaluations on four real datasets, where PeFAD outperforms existing state-of-the-art baselines by up to 28.74\%.

\begin{figure}[t]
	\centering
 \vspace{0.35cm}
		\includegraphics[width=0.99\linewidth]{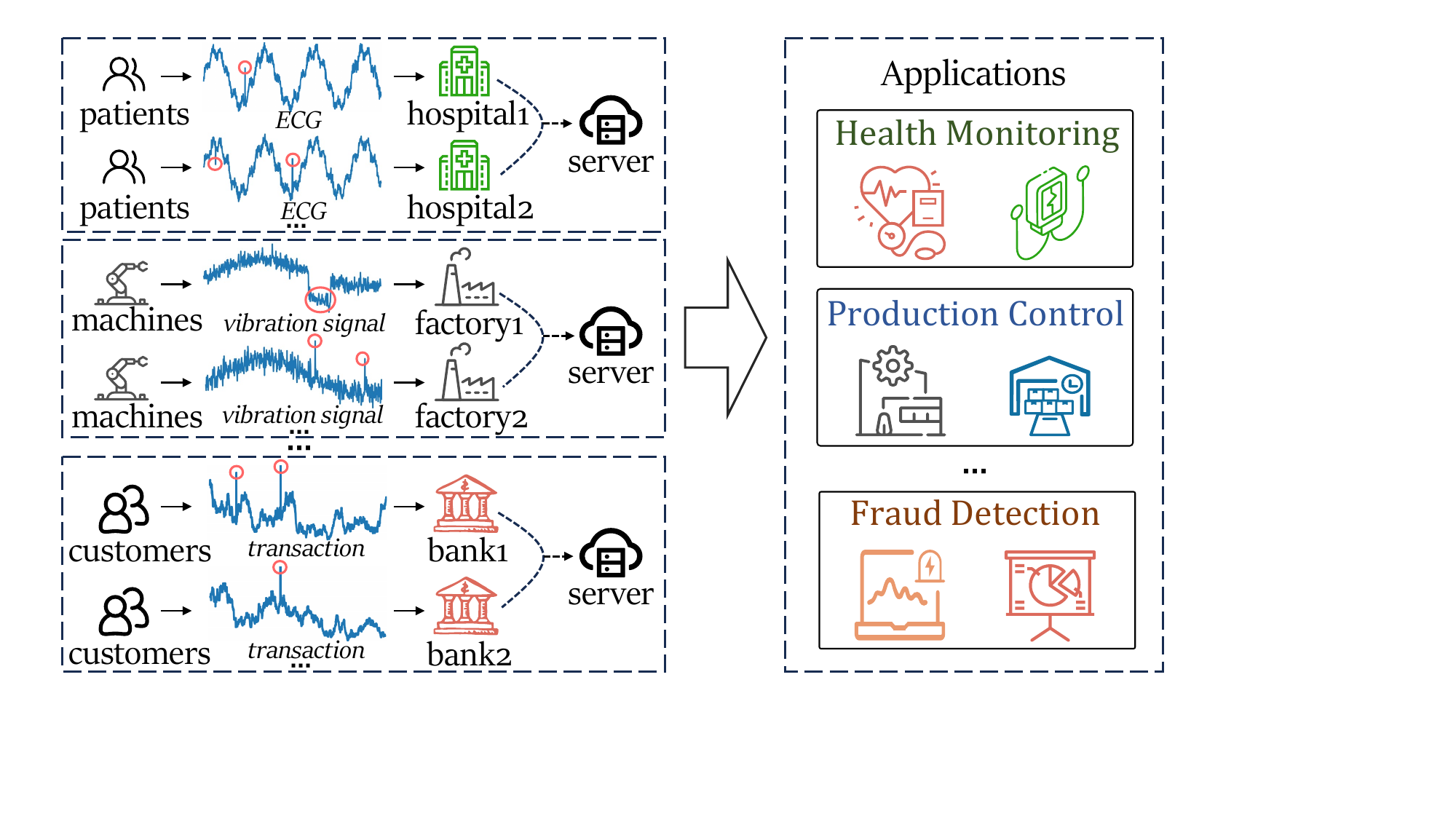}
            \captionsetup{font=small}
		\caption{Illustration of decentralized time series anomaly detection. "Red circles" denote anomaly points or anomalous patterns. In each scenario, data sharing between institutions is not allowed, and collaborative training is facilitated through server coordination.}
		\label{fig:illusion}%
\end{figure}
\end{abstract}

\ccsdesc[500]{Information systems~Data mining}
\ccsdesc[500]{Computing methodologies~Distributed algorithms}
\ccsdesc[500]{Computing methodologies~Anomaly detection}

\keywords{Time Series Anomaly Detection; Pre-trained Language Model; Federated Learning;}




\maketitle


\section{introduction}
With the increase of various sensors and mobile devices, massive volumes of time series data are being collected in a decentralized fashion, enabling various time series applications~\cite{wang2020deep, miao2022mba, miao2024unified, wu2023autocts+}, such as fault diagnosis~\cite{hsu2021multiple} and fraud detection~\cite{bolton2002statistical}. A fundamental aspect of these applications is time series anomaly detection~\cite{xu2022anomaly}, {as illustrated in Figure~\ref{fig:illusion}}, which aims to find unusual observations or trends in a time series that may indicate errors, or other abnormal situations requiring further investigations. 

Due to its significance, substantial research has been devoted to inventing effective time series anomaly detection models~\cite{bolton2002statistical, xu2022anomaly}, including approaches based on traditional statistics~\cite{liu2008isolation,tax2004support} and neural networks~\cite{xu2022anomaly}. 
Due to the difficulty in annotating anomalies, unsupervised methods become mainstream approaches, which can primarily be categorized into reconstruction-based~\cite{zhou2023one,xu2022anomaly} and prediction-based ~\cite{wu2021autoformer, zhou2021informer} approaches. The former identifies anomalies based on the reconstruction errors while the latter identifies anomalies based on the prediction errors.
In real-world scenarios, time series data is often generated by edge devices (e.g., sensors) that are distributed at different locations. However, most existing time series anomaly detection models generally require centralized training data, making them less effective in the decentralized scenarios. Due to the increasing concern on privacy protection, the data providers may not be willing to disclose their data. For instance, the credit agency Equifax experienced a data breach~\cite{zou2018ve} that exposed social security numbers and other sensitive data, significantly impacting individuals' financial security.
Therefore, decentralized time series anomaly detection has become a critical issue to enable privacy protection~\cite{mcmahan2017communication} and ensure data access restrictions~\cite{meng2021cross}.

Recently, Federated Learning (FL) has provided a solution for training a model with decentralized data distributed on multiple clients~\cite{mcmahan2017communication, yang2019federated}. FL is a machine learning setting where many clients collaboratively train a model under the orchestration of a central server while keeping data decentralized. In this study, we aim to develop a novel FL framework for unsupervised time series anomaly detection for bridging the gap between the decentralized data processing and the unsupervised time series anomaly detection. 

However, developing a federated learning-based time series anomaly detection model is non-trivial due to the following three challenges. 
First, it is challenging to deal with the data scarcity issue in the context of federated learning. Due to the limitation of data collection mechanisms (e.g., low sampling rates) and data privacy concerns, client-side local data can be very sparse, especially for the minority anomalous data. The performance of existing methods that rely on sufficient training data may degrade remarkably in the scenario of decentralized training data.
Second, existing unsupervised methods~\cite{xu2022anomaly,zhou2023one} often overlook the presence of anomalies during training. This may significantly disrupt the training process of both prediction and reconstruction-based methods, affecting their ability to accurately identify the anomalies~\cite{xu2022calibrated}. For instance, in reconstruction-based methods, if the masked time series fragments do not cover anomalous time points in training, the learned time series reconstruction model will be less sensitive to the anomalies~\cite{xiao2023imputation}. 
Third, it is also difficult to obtain a global model that generalizes well across all clients due to the heterogeneity of the local data.
The time series that are collected across different edge devices are typically heterogeneous and non-identical distributed ~\cite{zhang2023navigating}.
It is non-trivial for a FL model to achieve an optimal global model by simply aggregating local models due to the distribution drift across different local time series datasets.  

To address the above challenges, this paper proposes a \underline{P}arameter-\underline{e}fficient \underline{F}ederated time series \underline{A}nomaly \underline{D}etection framework named PeFAD. PeFAD adopts a horizontal federated learning schema, where many clients collaboratively train a global model by using the local training data under the orchestration of a central server. PeFAD contains two major modules: the PLM-based local training module and the parameter-efficient federated training module. The PLM-based local training module employs the pre-trained language model (PLM) for each client, which features an anomaly-driven mask selection strategy and a privacy-preserving shared dataset synthesis mechanism.
We adopt the PLM as the body of the local model of clients because its cross-modality knowledge transfer capability~\cite{lu2022frozen,zhou2023one, liu2024spatial} can effectively address the challenge of data scarcity. Specifically, we aim to leverage the generic knowledge and the contextual understanding capability of PLM to help discern the time series patterns and anomalies. To reduce the computation and communication overhead of PLM, we propose a parameter-efficient federated training module. The clients only need to fine-tune small-scale parameters and then transfer them to the server. 
In order to mitigate the impact of anomalies during training, we propose a novel anomaly-driven mask selection strategy to first identify anomalies during training, and then assign them larger weights to be selected for masking. 
To alleviate the data heterogeneity across clients, we propose a privacy-preserving shared dataset synthesis mechanism. To be specific, each client first utilizes a variational autoencoder to synthesize privacy-preserving time series, and the synthesized data are pooled together to form a dataset shared by all clients. Then knowledge distillation is performed between local and global models with the shared dataset to achieve a more consistent model update between the clients.

Our primary contributions are summarized as follows.
\begin{itemize}
\item To the best of our knowledge, this is the first PLM-based federated framework for unsupervised time series anomaly detection. 
To reduce the computation and communication costs, we propose a parameter-efficient federated  training module.


\item To alleviate the impact of anomalies during training, an anomaly-driven mask selection strategy is proposed, which enhances the model's adaptability towards change points, thereby improving the robustness of anomaly detection.

\item To deal with the data heterogeneity across clients, a novel privacy-preserving shared dataset synthesis mechanism and a knowledge distillation method are both proposed to ensure a more consistent model updating between clients.

\item We conduct extensive evaluations on four popular time series datasets. The result demonstrates that the proposed PeFAD significantly outperforms existing SOTA baselines in both centralized and federated settings.
\end{itemize}





The remainder of this paper is organized as follows. Section~\ref{sec:RELA} reviews related work and analyzes the limitations of existing work. Section~\ref{sec:PRELIM} introduces preliminary concepts and the federated time series anomaly detection problem. We then 
 present our solutions in Section~\ref{sec:METHOD}, followed by the experimental evaluation in Section~\ref{sec:EXPS}. Section~\ref{sec:DISSCUSS} discuss the results to the motivation of the paper, and Section~\ref{sec:CONCLU} concludes the paper.
\section{Related work}\label{sec:RELA}

\subsection{Time Series Anomaly Detection}
Time series anomaly detection aims to identify unusual patterns or outliers within time series, which plays a crucial role in various real-world applications~\cite{shang2016intrusion, xu2022anomaly}. 
Traditionally, time series anomaly detection methods are mostly based on conventional machine learning models such as support vector machine (SVM)~\cite{shang2016intrusion} and isolation forest~\cite{liu2008isolation}. 
The major limitation of the above methods is that the complex temporal correlations of time series are hard to be captured due to their limited learning capability.
Recently, with the advances in deep learning techniques, deep neural network models have been widely used for time series anomaly detection, which can be categorized into supervised and unsupervised methods. 
Supervised methods~\cite{pang2021toward} are trained on labeled data to identify deviations from normal patterns in time series.
Unsupervised methods~\cite{xu2022anomaly,zhou2023one} often calculate an anomaly score to measure the difference between the original time series and the reconstructed or predicted time series.
The unsupervised methods can learn the intrinsic structure and patterns of time series beyond the labels.
Nevertheless, existing time series anomaly detection methods are mostly trained with centralized data and are computational heavily, limiting their usage on resource-constrained edge devices. 
\begin{figure*}[t]
    \centering
    \includegraphics[width=0.95\linewidth]{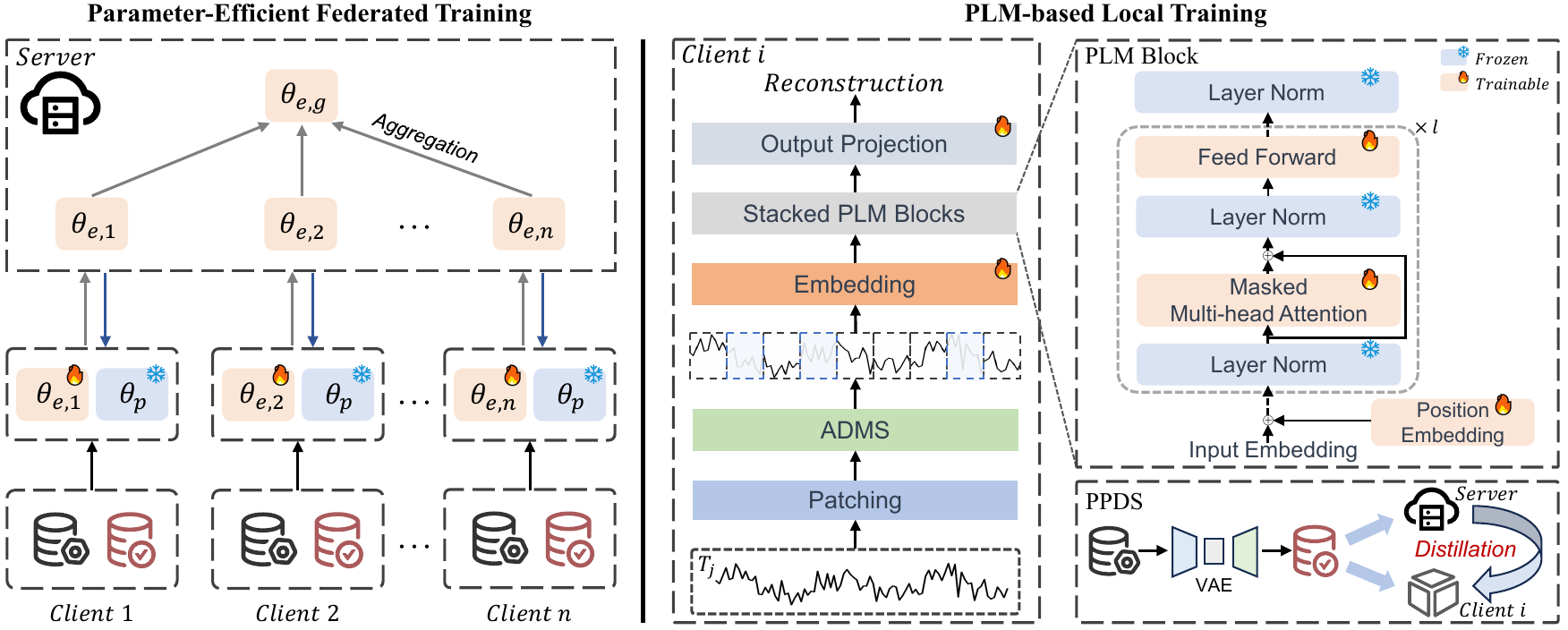}
    \caption{PeFAD framework overview. PeFAD consists of PLM-based local training and parameter-efficient federated training. }
    \label{fig:1}
\end{figure*}

\subsection{Federated Learning}
Federated learning (FL) is a machine learning approach in which many clients (commonly referred to as edge devices) collaboratively train a model using decentralized data
~\cite{mcmahan2017communication,fan2022improving,liu2022vertical,saha2021federated, liu2024icde}. Typically, FL can be categorized into horizontal federated learning, vertical federated learning, and federated transfer learning based on the overlap of data features and sample space among clients~\cite{mcmahan2017communication}. Horizontal FL~\cite{fan2022improving} is defined as the case where datasets on different clients share the same feature space but have different sample space, while vertical FL~\cite{liu2022vertical} is the opposite case. In federated transfer learning~\cite{saha2021federated}, the sample space and feature space between cross-client data are virtually non-overlapping.
In this study, we consider time series anomaly detection based on horizontal FL.

Recently, FL has been applied to time series with the concern of privacy protection, such as time series forecasting~\cite{meng2021cross} and anomaly detection~\cite{liu2022fedtadbench}. 
However, existing research lacks an in-depth exploration on how to use pre-trained language models for time series anomaly detection in a federated setting, leaving a significant gap in the existing literature. This gap can be attributed to the inherent complexities associated with reconciling domain differences and task variations within the context of federated learning when applying pre-trained language models. 

\section{Problem definition}\label{sec:PRELIM}
We first present the necessary preliminaries and then define the problem addressed.
To make notations consistent, we use \textbf{bold} letters to denote matrices and vectors.

\begin{definition}[Time Series]
    A time series ${T} = \langle \bm{t}_1, \bm{t}_2, \cdots, \bm{t}_m \rangle$ is a time ordered sequence of $m$ observations, where each observation $\bm{t}_i \in \mathbb{R}^D$ is a $D$-dimensional vector. If $D=1$, $T$ is univariate, and if $D>1$, $T$ is multivariate.
\end{definition}


\textbf{Federated Time Series Anomaly Detection.} Given a server $\mathcal{S}$ and $\mathcal{N}$ clients (e.g., sensors) with their local time series datasets $\mathcal{D} = \left\{\mathcal{T}_1, \mathcal{T}_2, \cdots, \mathcal{T}_{\mathcal{N}}\right\}$, each dataset $\mathcal{T}_i$ is a set of time series, i.e., $\mathcal{T}_i = \left\{{T_1^i}, {T_2^i}, \cdots, {T_n^i} \right\}$. We aim to learn a shared global function $\mathcal{F}(\theta)$ that can detect anomalies in time series across different clients. The optimal global model parameters $\theta^*_g$ is obtained as follows:
 
\begin{equation}
\begin{split}
&\theta^*_g = \arg \min_{\theta_g}  \sum_{i\in \mathbb{C}} \frac{|\mathcal{T}_i|}{\sum_{j\in \mathbb{C}}|\mathcal{T}_j|}  \mathbb{E}_{\mathcal{T}_i}[\mathcal{L}(\theta_g;\mathcal{T}_i)],
\label{eq:4}
\end{split}
\end{equation}
where $\mathcal{L}(\theta_g;\mathcal{T}_i)$ denotes the loss function for client $i$, and $\theta_g$ denotes parameters of the global model. $\mathbb{C}$ denotes the set of clients.

In client $i$, given a time series ${T^i} = \langle \bm{t}^i_{1}, \bm{t}^i_{2}, \cdots, \bm{t}^i_{m} \rangle$, we aim at computing an outlier score $OS(\bm{t}^i_j)$ for each time point $j$. A higher $OS(\bm{t}^i_j)$ means it is more likely that $\bm{t}^i_j$ is an outlier. The outlier score can be formulated as follows:
\begin{equation}
    OS(\bm{t}^i_j) = 
    |\bm{t}^i_j - \bm{\hat{t}}^i_j|; \, s.t. \bm{\hat{{t}}}^i_j = \mathcal{F}(\theta^*_g,\bm{t}^i_j), 
\end{equation}
where $\bm{\hat{t}}^i_j$ is the reconstructed value of $\bm{t}^i_j$. We consider the top $r\%$ of $OS(\bm{t}^i_j)$ as anomalies, where $r$ is a threshold.

\section{METHODOLOGY}\label{sec:METHOD}
Figure~\ref{fig:1} shows the framework overview of the proposed PeFAD. As shown in the figure, PeFAD consists of two major modules: PLM-based local training (right part of the figure) and parameter-efficient federated training (left part of the figure). Specifically, in PLM-based local training module, the client first uses a patching mechanism and the anomaly-driven mask selection strategy (ADMS) to preprocess the local time series, such that the model can better understand the complex patterns of time series. Then the preprocessed data is input into the PLM-based local model for training. Specifically, the preprocessed data undergoes embedding layer, the stacked PLM blocks, and the output projection layers to finally output the reconstructed time series. Based on the reconstructed data, the client identifies the anomalous points by calculating the reconstruction error.
Furthermore, a privacy-preserving shared dataset synthesis mechanism (PPDS, lower right part of the figure) is utilized to alleviate data heterogeneity across clients through knowledge distillation. To reduce computation and communication cost, we also propose a parameter-efficient federated training module.
Next, we will provide the technical details of each module, respectively.

\subsection{PLM-based Local Training}\label{sec:4.2}

To better capture local temporal information, the client divides the local time series into non-overlapping patches~\cite{nie2022time}. Specifically, we aggregate adjacent time steps to create patch-based time series. This application of patching allows for a substantial extension of the input historical time horizon while keeping the token length consistent and minimizing information redundancy for transformer models.
Then, we select a certain proportion of these patches for masking using an anomaly-driven mask selection strategy. 

\subsubsection{{\textbf{Anomaly-Driven Mask Selection.}}}
Existing reconstruction based methods~\cite{xu2022anomaly,zhou2023one,wu2022timesnet} generally neglect the anomalies in the training data, which may disrupt mask reconstruction. 
For instance, if normal points are masked while anomalous points are utilized as observations to reconstruct the masked time series fragments, it may result in large reconstruction errors~\cite{xu2022calibrated}.
To address this issue, we propose the anomaly-driven mask selection strategy to first identify the anomalies, and then assign them larger weights to be chosen for masking. 
The module combines the analysis on intra- and inter-patch variability to calculate the anomaly score of patches, capturing both patch-specific deviations and the contextual evolution of patterns over time.

\textbf{Intra-patch Decomposition.} To capture the intrinsic characteristics of the $i$-th patch (denoted as $\bm{{P}}_{i}$), we utilize time series decomposition technique~\cite{hassani2007singular}. Specifically, we decompose each patch into $M$ components, as formulated in Eq.~\eqref{eq:depcompose}, and extract residual components to calculate the intra-anomaly score of patches.

\begin{equation}
    \bm{{P}}_{i} = \sum_{j=1}^{M} a_j \bm{g}_j + \varepsilon,\ \ s.t.\ a_j \geq 0,\ \forall j,\  \sum_{j=1}^{M} a_j = 1,
    \label{eq:depcompose}
\end{equation}
where $\bm{g}_j$ denotes the $j$-th component, $a_j$ is the coefficient for $j$-th component, and $\varepsilon$ denotes the noise term.

Specifically, we use Singular Spectrum Analysis (SSA)~\cite{hassani2007singular} to decompose patches. In SSA, patch $\bm{{P}}_i$ is first transformed into a Hankel matrix $\bm{\mathcal{P}}_i$ through embedding, and then Singular Value Decomposition (SVD) is applied to the matrix, decomposing $\bm{\mathcal{P}}_i$ into the product of three matrices:  $\bm{\mathcal{P}}_i = \bm{U} \bm{\Sigma} \bm{V}^T$, 
where $\bm{U}$ and $\bm{V}$ denote the left and right singular vector matrices, respectively, and $\bm{\Sigma}$ denotes the diagonal matrix of singular values.
Then, the original patch is reconstructed by 
\begin{equation}
\begin{split}
    &\bm{{\mathcal{P}}}_{i} = \sum_{k=1}^{K} \bm{\sigma}_k \bm{u}_k \bm{v}_k^T = \sum_{k=1}^{K} \bm{{\mathcal{P}}}_{i,k},
    \label{eq:Ht}
\end{split}
\end{equation}
where $K$ denotes number of non-zero eigenvalues of $\bm{{\mathcal{P}}}_{i}$. $\bm{\sigma}_k$ is the $k$-th singular value, $\bm{u}_k$ is the $k$-th left singular vector, and $\bm{v}_k$ is the $k$-th right singular vector.

Matrix $\bm{{\mathcal{P}}}_{i}$ constitutes the main structure of the original patches. For instance, the trend, seasonal, and residual components correspond to the low, mid, and high frequency components of matrix $\bm{{\mathcal{P}}}_{i}$. We can obtain these components by filtering.
Residuals often contain anomalies in the time series~\cite{schmidl2022anomaly}. Therefore, we extract the residual component after decomposition, and calculate the mean of the residual components as the residual value $\mathcal{R}_i$, as formulated in Eq.~\eqref{eq:Ri}. A higher residual value indicates a larger likelihood to be an anomaly. We then normalize $\mathcal{R}_i$ to calculate the anomaly score $\mathcal{R}_i^{'}$ for the $i$-th patch.
\begin{equation}
    \mathcal{R}_i = mean(\sum_{k\in K_N}\bm{\mathcal{P}}_{i,k}) =  mean(\sum_{k\in K_N}{\sigma}_k {u}_k {v}_k^T),
\label{eq:Ri}
\end{equation}
where subscript $k$ denotes the $k$-th value of the matrix, and $K_N$ denotes the set of singular values associated with residual components obtained by filtering.

\textbf{Inter-patch Similarity Assessment.}
The inter-patch similarity assessment provides insights into the dynamic evolution of patterns patches. 
Assuming $\bm{\mathcal{A}}_i$ is the vector of patch $i$, we calculate the cosine similarity between the $i$-th and ($i$-$1$)-th patches.
\begin{equation}
    \mathcal{C}_i = \frac{\bm{\mathcal{A}}_{i} \cdot \bm{\mathcal{A}}_{i-1}}{\|\bm{\mathcal{A}}_{i}\| \cdot \|\bm{\mathcal{A}}_{i-1}\|}.
    \label{eq:cos}
\end{equation}


The cosine similarity ranges from -1 to 1, and a larger value indicates a higher similarity between patches. Patches with lower similarity to the previous patches are more likely to be anomalous, so we alter the monotonicity and normalize $C_i$ to calculate the anomaly score $C_i^{'}$ for the $i$-th patch.

\textbf{Anomaly Score of Patches.}
We synthesize the intra-patch time series decomposition and the inter-patch similarity assessment to obtain a final anomaly score for patch $i$ as follows:
\begin{equation}
    Score_i = \beta * \mathcal{R}_i^{'} + (1-\beta) * \mathcal{C}_i^{'}\label{eq:}.
\end{equation}

The patches whose anomaly scores surpass a predefined threshold are considered as anomalies and are assigned larger weights to be chosen for masking. Since the masked patches are more emphasized by the model, the anomaly-driven mask selection strategy can enhances the model’s adaptability towards change points, thus improving the robustness of anomaly detection.

\subsubsection{\textbf{Privacy-Preserving Shared Dataset Synthesis}}\label{sec:4.3}
In federated learning, clients may have different data distributions and features, posing a data heterogeneity challenge that makes the generalization of the aggregated model difficult. 
To address this issue, we propose a privacy-preserving shared dataset synthesis scheme coupled with knowledge distillation.

\textbf{Privacy-Preserving Shared Dataset Synthesis.} 
Recent works have demonstrated that reducing mutual information can facilitate privacy protection in dataset generating~\cite{yang2023fedfed}.
Inspired by this idea, we employ a constrained mutual information approach to obtain synthetic data for preserving the privacy of local data. Specifically, Client $i$ trains a variational autoencoder (VAE) model to synthesize time series $\mathcal{T}_{s,i}$ from the local time series $\mathcal{T}_i$. The mutual information $I(\mathcal{T}_i;\mathcal{T}_{s,i})$ measures the extent to which $\mathcal{T}_{s,i}$ reveals $\mathcal{T}_i$. 
Through constraining   $I(\mathcal{T}_i;\mathcal{T}_{s,i})$, the likelihood of inferring $\mathcal{T}_i$ from $\mathcal{T}_{s,i}$ has been reduced, thereby better protecting data privacy and facilitating the synthesis of privacy-preserving time series. 
\begin{equation}
 I(\mathcal{T}_i;\mathcal{T}_{s,i}) = \sum_{x \in \mathcal{T}_i} \sum_{y \in \mathcal{T}_{s,i}} p(x, y) \log \left( \frac{p(x, y)}{p(x)p(y)} \right), \label{eq:6}
\end{equation}
where \(p(x, y)\) denotes the joint probability distribution, with \(p(x)\) and \(p(y)\) as the marginal probabilities of $x$ and $y$, respectively. 

In order to ensure the validity of the synthesized time series, we introduce a constraint to maintain the distribution similarity between the synthesized and the original time series. We use Wasserstein distance to quantify this distribution similarity~\cite{ruschendorf1985wasserstein}. A smaller Wasserstein distance indicates a lower cost of transforming from one distribution to another, implying that the two distributions are more similar. Given two time series $X$ = $\langle \bm{x}_1, \bm{x}_2, \ldots, \bm{x}_m \rangle$ and $Y$ = $\langle \bm{y}_1, \bm{y}_2, \ldots, \bm{y}_n \rangle$, and their cumulative distribution functions \(F_X\) and \(F_Y\), the Wasserstein distance can be obtained as follows,
\begin{equation}
\begin{split}
 &\ F_X(x) = \frac{1}{m} \sum_{i=1}^{m} 1_{\{\bm{x}_i \leq x\}},\ \ F_Y(y) = \frac{1}{n} \sum_{j=1}^{n} 1_{\{\bm{y}_j \leq y\}},\\
    &W(X, Y) = \inf_{\gamma \in \Gamma(F_X, F_Y)} \int_{-\infty}^{\infty} |F_X(x) - F_Y(y)| \, d\gamma(x, y),
     \label{eq:wd}
\end{split}
\end{equation}
where $\gamma$ denotes the joint distributions between $F_X$ and $F_Y$, and 
$\Gamma(F_X, F_Y)$ denotes the set of all joint distributions with the marginal distributions $F_X$ and $F_Y$. 



We use VAE to synthesize time series, which consists of an encoder and a decoder. The encoder first encodes the input time series as a feature representation, and the decoder then attempts to generate a synthesized time series based on the representation. The raw data privacy and the synthesized data validity are guaranteed by constraining mutual information and Wasserstein distance, respectively. The loss function for VAE is given by
\begin{equation}
\begin{split}
    &\min_{\mathcal{T}_{s,i}} \,  \mathcal{L}_{vae} + \alpha_1 \cdot W(\mathcal{T}_i, \mathcal{T}_{s,i})  + \alpha_2 \cdot I(\mathcal{T}_i;\mathcal{T}_{s,i}), \\
   &\mathcal{L}_{vae} = -\mathbb{E}_{q(\bm{z}|\bm{x})}[\log p(\bm{x}|\bm{z})] + KL[q(\bm{z}|\bm{x}) \, || \, p(\bm{z})],
 \label{eq:vae_loss}
 \end{split}
\end{equation}
where $\mathcal{L}_{vae}$ denotes the base loss function of VAE. $\bm{x}$ and $\bm{z}$ denote the input and latent vectors, respectively. $q(\bm{z}|\bm{x})$ and $p(\bm{x}|\bm{z})$ denote the output distributions of the encoder and decoder, respectively. $KL(\cdot)$ denotes the Kullback-Leibler divergence ~\cite{van2014renyi}, which can be calculated as follows:
\begin{equation}
{KL}\left(q(\bm{z}|\bm{x}) \, || \, p(\bm{z})\right) = \frac{1}{2} \sum_{i} \left( \sigma_i^2 + \mu_i^2 - \log(\sigma_i^2) - 1 \right),
 \label{eq:kl}
\end{equation}
where both $q(\bm{z}|\bm{x})$ and $p(\bm{z})$ are assumed to follow multivariate Gaussian distributions. $\mu_i$ and $\sigma_i$ are the mean and standard deviation of the Gaussian distribution.

Then, the server integrates the synthesized time series from clients to form a shared dataset $\mathcal{D}_{sh}$. Note that time series synthesis is a one-time offline process before local training. 
\begin{equation}
\mathcal{D}_{sh} = \bigcup_{i\in \mathbb{C}} \mathcal{T}_{s,i} = \langle \mathcal{T}_{s,1}, \mathcal{T}_{s,2} ,..., \mathcal{T}_{s,\mathcal{N}} \rangle.
\label{eq:10}
\end{equation}


\textbf{Knowledge Distillation.} 
We further perform knowledge distillation from the global model to the client models using the shared dataset to reduce the data heterogeneity across clients.
Specifically, we first obtain the learned representations of the local and global models on the shared dataset separately, and then calculate the difference between the two representations. 
We use the consistency loss to measure this difference. Through reducing this discrepancy, the model can achieve more consistent client updates, thereby improving the performance and stability of the aggregated global model.
The consistency loss is introduced as a regularization term to the local loss function as follows,
\begin{equation}
    \mathcal{L}(\theta_i;\mathcal{T}_i) = \underbrace{\frac{1}{n}\sum_{j=1}^{n}|{\hat{T}^i_j}-{T^i_j}|^2}_{\text{Reconstruction Loss}} +\ \lambda \cdot \underbrace{\| \mathcal{F}(\theta_i,\mathcal{D}_{sh})-\mathcal{F}(\theta_g,\mathcal{D}_{sh}) \|}_{\text{Consistency Loss}},
      \label{eq:loss} \tag{$13$}
\end{equation}
where $\hat{T}^i_j$ and $T^i_j$ denote the reconstructed and real values of $j$-th time series of client $i$, respectively. $\theta_i$ and $\theta_g$ represent the parameters of the $i$-th local and global model, respectively.
$\lambda$ is a parameter to trade off the two loss terms.

\subsection{Parameter-Efficient Federated Training}
\label{sec:4.4}

As a horizontal FL framework, PeFAD comprises a central server and several clients. 
The local model of each client consists of an input embedding layer, the stacked pre-trained language model (PLM) blocks, and an output projection layer, as illustrated on the right part of Figure~\ref{fig:1}. GPT2 is used as the PLM~\cite{radford2019language}. 
We first adopt several linear layers to embed the raw time series data into the feature representations required by the PLM. The output of PLM undergoes a fully connected layer to convert the output dimension of GPT2 to the dimension that the data reconstruction model needs~\cite{zhou2023one}.

We divide the model parameters into trainable parameters $\theta_e$ and frozen parameters $\theta_p$, i.e. $\theta = (\theta_e, \theta_p)$.
We frozen the majority of parameters in the PLM, that is, $|\theta_e| \ll |\theta|$. 
Specifically, the frozen parameters include the layer normalization blocks and the first $n$ layers ($n\geq5$).
We choose to freeze the majority of the parameters of the PLM during fine-tuning as they encapsulate most of the generic knowledge learned from pre-training phase. 
To enhance downstream time series anomaly detection tasks with minimal effort,
we fine-tune the input-output layers and certain parts of the last one or three layers of the PLM, including the attention layer, the feed-forward layer, and positional embedding, as they contain task-specific information and adjust them allows the model to adapt to the nuances of the target domain or task.
The process of parameter-efficient federated training module is given in Algorithm~\ref{algo:1}. 

\textbf{Training on Server Side.} The server first sends trainable parameters $\theta_e$ to the clients for initialization (Lines~\ref{line:5}). Then, client $i$ updates $\theta_{e,i}$ through local training (Line~\ref{line:6}). Finally, server receives parameters from all clients and aggregates them to get updated parameters $\theta_{e,g}$ (Lines~\ref{line:7}--~\ref{line:8}).


\textbf{Local Training on Client Side.} After the clients receive $\theta_{e,g}$ from the server, they assemble the whole PLM model with trainable parameters $\theta_{e,i}$ and frozen parameters $\theta_{p}$ (Line~\ref{line:10}). The $i$-th local model updates its parameters $\theta_{e,i}$ by gradient descent (Lines~\ref{line:11}--~\ref{line:14}). After the local training is completed, client sends $\theta_{e,i}$ to the server for aggregation (Line~\ref{line:15}).

The training process described above is repeated until PeFAD converges according to Eq.~\eqref{eq:4}. 
\begin{algorithm}[h]
  \caption{Parameter-Efficient Federated Training}
  \label{algo:1}
  \SetNlSty{mysty}{\normalfont\bfseries}{:}

  \SetAlgoNlRelativeSize{-2}

  \SetKwInput{Parameters}{Input}
  \SetKwInput{Initialize}{Initialization}
  \SetKwInput{Output}{Output}
  \SetKwFunction{ServerGlobalAggregation}{\textbf{Server Execute }}
  \SetKwFunction{ClientLocalTraining}{\textbf{Client Update}}
  
  \Parameters{model parameters $(\theta_e,\theta_p)$; clients set $\mathbb{C}$; global and local epoch number: $T_g$ and $T_l$; learning rate $\eta$; weight coefficient $\lambda$; dataset $\mathcal{D}=\{\mathcal{T}_1, \mathcal{T}_2,..., \mathcal{T}_\mathcal{N}\}$; local dataset $\mathcal{T}_i = \left\{{T_1^i}, {T_2^i}, \cdots, {T_n^i} \right\}$;}
  \Output{Trained global model $\theta_g$.}
  \ServerGlobalAggregation:{\label{line:1}\\
  \ \ Initialize the trainable parameters $\theta_{e,g}^0$;\\
   \ \  \For{global round $t_g = 1$ to $T_g$}{

      \For{each client $i\in \mathbb{C}$  in parallel}
      {
        Initialize client model $\theta_{e,i}^{{t_g}-1} = \theta_{e,g}^{{t_g}-1}$; \label{line:5}\\
        \ClientLocalTraining{$i$, $\theta_{e,i}^{{t_g}-1}$}\; \label{line:6}
      }
      Receive $\theta_{e,i}^{{t_g}}$ from all clients in $\mathbb{C}$ ; \label{line:7} \\
      Update $\theta_{e,g}^{{t_g}}$ by:\ \ $\theta_{e,g}^{t_g} = \sum_{i\in \mathbb{C}} \frac{|\mathcal{T}_i|}{\sum_{j \in \mathbb{C}}|\mathcal{T}_j|} \cdot \theta_{e,i}^{t_g} $\;\label{line:8}
    }
  }
  
  \vspace{1em} 

  \ClientLocalTraining($i$, $\theta_{e,i}^{t_g-1}$):{ \label{line:9} \\
  \ \   $\theta_{i}^{t_g-1} \leftarrow$ (assemble $\theta_{e,i}^{t_g-1}$ and $\theta_p$)\; \label{line:10}
  \ \   \For{local round $t_l = 1$ to $T_l$}{ \label{line:11}
         $\mathcal{L} = \frac{1}{n}\sum_{j=1}^{n}|{\hat{T}^i_j}-{T^i_j}|^2\ + $\\ \ \ \ \ \ \ \ \ 
         $\lambda \cdot\| \mathcal{F}(\theta_i^{({t_g-1},t_l)},\mathcal{D}_{sh})\ - 
       \mathcal{F}(\theta_g^{{t_g-1}},\mathcal{D}_{sh}) \| $;\\
        $\theta_{e,i}^{{(t_g,t_l)}} \leftarrow \theta_{e,i}^{({t_g-1},t_l)} - \eta \cdot \nabla \theta_{e,i}^{({t_g-1},t_l)} \mathcal{L}_i$; \label{line:14}
    }
  \ \   Send $\theta_{e,i}^{{t_g}}$ to the server\;\label{line:15}
  }
  \Return $\theta_g$
\end{algorithm}

\subsection{Overall Objective}
In this section, we give the overall objective of the proposed method. 
For client $i$, it updates the local trainable model parameters by optimizing the loss function $\mathcal{L}$, and sends the trainable parameters to the server.
\begin{equation}
\mathcal{L}(\theta_i;\mathcal{T}_i) = \frac{1}{n}\sum_{j=1}^{n}|{\hat{T}^i_j}-{T^i_j}|^2+ \lambda \cdot\| \mathcal{F}(\theta_i,\mathcal{D}_{sh}) -  \mathcal{F}(\theta_g,\mathcal{D}_{sh}) \|,
\label{eq:loss-1} 
\end{equation}
where ${\hat{T}^i_j}$ and ${T^i_j}$ denote the reconstructed and real values of $j$-th time series of client $i$, respectively. $\theta_i$ and $\theta_g$ represent the parameters of the $i$-th local model and global model, respectively, composed of trainable parameters $\theta_e$ and frozen parameters $\theta_p$. 


The server aggregates trainable parameters across clients within the global iteration rounds to obtain the global model.
\begin{equation}
    \theta_{e,g}^t = \sum_{i \in \mathbb{C}} \frac{|\mathcal{T}_i|}{\sum_{j \in \mathbb{C}}|\mathcal{T}_j|} \cdot \theta_{e,i}^t . 
    \label{eq:aggre}
\end{equation}

The time series anomaly detection for each client is achieved by leveraging the aggregated global model. To detect anomalies, we input the testing time series into the local model to obtain its reconstructed values at all time points. The anomaly score at time point $k$ is computed based on the reconstruction error $re$ as follows,
\begin{equation}
    re = |\bm{t}_k - \bm{\hat{t}}_k|,
\end{equation}
where $\bm{t}_k$ and $\bm{\hat{t}}_k$ are the real and reconstructed values at time point $k$, respectively.



\captionsetup{} %
\begin{table*}[t]
  \caption{{Quantitative results for various methods on four datasets. P, R, AUC and F1 denote Precision, Recall, AUC-ROC and F1-Score as \% , respectively. "Central." represents centralized.}}
  \label{tab:main_res}
  \resizebox{1\textwidth}{!}{
  \renewcommand{\arraystretch}{1.1} 
  \begin{tabular}{p{0.95cm}|c|cccc|cccc|cccc|cccc}
  
  \toprule
    \multirow{2}{*}{} & 
    \multirow{2}{*}{Methods} & \multicolumn{4}{c|}{SMD} & \multicolumn{4}{c|}{PSM}  & \multicolumn{4}{c|}{SWaT}& \multicolumn{4}{c}{MSL} \\
    \cmidrule(lr){3-6} \cmidrule(lr){7-10} \cmidrule(lr){11-14} \cmidrule(lr){15-18}

  & & {P} & {R} & {AUC} & {F1} & {P} & {R} & {AUC} & {F1} & {P} & {R} & {AUC} & {F1} & {P} & {R} & {AUC} & {F1} \\
    \midrule
    \centering
    \multirow{12}{*}{{Central.}}
      &{OCSVM} & {4.87} & 23.44 & 49.02 & 8.01 & 24.11 & 69.49 & 31.96 & 35.80 & 77.91 & 64.18 & 19.39 & 70.38 & 19.01 & 19.86 & 52.25 & 19.42 \\
     &{IF} & 9.02 & 39.00 & 32.84 & 14.66 & 24.25 & 52.42 & 42.47 & 33.16 & 75.76 & 62.40 & 18.78 & 68.44 & 9.55 & 58.57 & 41.58 & 16.42 \\
     &{LOF} & 8.19 & 19.72 & 44.93 & 11.58 & 34.27 & 12.35 & 48.38 & 18.15 & 14.01 & 11.54 & 49.12 & 12.66 & 13.06 & 12.92 & 48.37 & 13.25  \\
     &{MTGFLOW} & 91.21 & 67.22 & 83.47 & 77.40 & 99.71 & 86.66 & 93.28 & 92.73 & 96.61 & 83.56 & 91.58 & 89.61 & 97.25 & 63.40 & 81.59 & 76.76  \\
     &{GANF} & 88.31 & 68.31 & 84.46 & 77.67 & 98.62 & 82.01 & 90.79 & 89.55 & 96.36 & 79.01 & 89.30 & 86.83 & 97.15 & 63.20 & 81.49 & 76.58  \\
     &{Autoformer} & 78.45 & 65.10 & 82.16 & 71.15 & 99.94 & 79.06 & 89.52 & 88.28 & 99.90 & 65.55 & 82.77 & 79.16 & 76.93 & 76.50 & 86.90 & 76.71 \\
     &{Informer} & 90.28 & 75.24 & 87.14 & 82.08 & 97.29 & 80.59 & 89.86 & 88.15 & 99.83 & 67.87 & 83.93 & 80.80 & 79.79 & 74.73 & 86.25 & 77.18 \\
     &{FEDformer} & 76.78 & 59.72 & 79.47 & 67.19 & 99.98 & 81.69 & 90.84 & 89.91 & 99.94 & 65.61 & 82.80 & 79.22 & 90.61 & 69.02 & 84.09 & 78.35 \\
     &{TimesNet} & 88.00 & 81.44 & 90.48 & 84.59 & 97.32 & 96.62 & {97.76} & 96.97 & 85.50 & 93.69 & 95.75 & {89.41} & 88.78 & 73.61 & 86.26 & 80.48  \\
     &AT & 90.34 & 82.34 & {90.98} & {86.16} & 95.70 & 95.34 & 96.85 & 95.52 & 76.79 & 80.02 & 88.34 & 78.37 & 69.14 & 86.48 & \textcolor{red}{90.97} & 76.85  \\
     &{FPT} & 87.60 & 80.79 & 90.15 & 84.06 & 98.36 & 95.82 & 97.60 & {97.07} & 79.80 & 97.04 & {96.09} & 87.58 & 81.10 & 80.35 & {89.07} & {80.72}  \\
    &{\ \ PeFAD$_c$} & 87.93 & 94.37 & \textcolor{red}{97.00} & \textcolor{red}{90.72} & 97.99 & 97.47 & \textcolor{red}{98.37} & \textcolor{red}{97.72} & 91.19 & 94.91 & \textcolor{red}{96.82} & \textcolor{red}{93.01} & 80.87 & 82.73 & {90.22} & \textcolor{red}{81.79} \\
     \midrule
     \centering
    \multirow{8}{*}{{FL}} 
    &{Autoformer$_{fl}$} & 74.92 & 82.30 &90.74 & 77.23 & 97.77 & 78.88 & 89.12 & 86.64 & 95.04 & 66.68 & 83.26 & 77.59  & 84.09 & 65.57 & 82.42 & 72.66\\
     &{Informer$_{fl}$} & 77.44 & 91.18 & \underline{95.18} & 83.08 & 77.98 & 59.58 & 72.20 & 64.11 & 39.84 & 27.20 & 59.42 & 30.49 & 80.34 & 67.90 & 83.52 & 72.12\\
      &{FEDformer$_{fl}$} & 76.64 & 89.58 & 94.37 & 81.66 & 76.69 & 58.54 & 71.65 & 62.64 & 40.23 & 29.40 & 60.52 & 32.55 & 79.16 & 66.95 & 83.02 & 71.36\\
    &{TimesNet$_{fl}$} & 86.36 & 85.30 & 92.44 & 84.97 & 98.30 & 89.84 & 94.64 & 93.75 & 88.19 & 84.61 & 91.77 & 86.22 & 70.69 & 73.69 & \underline{85.80} & 71.53\\
    &{AT$_{fl}$} & 87.02 & 83.57 & 91.62 & 84.63 & 97.29 & 80.02 & 89.62 & 87.07 & 49.96 & 41.77 & 70.88 & 45.50 & 81.77 & 69.40 & 83.96 & \underline{73.93}  \\
     &{FPT$_{fl}$} & 84.93 & 80.08 & 89.85 & 81.49 & 98.56 & 91.78 & \underline{95.66} & \underline{94.92} & 88.07 & 85.66 & \underline{92.28} & \underline{86.74} & 70.90 & 73.25 & 85.52 & 71.85  \\
    &FedTADBench & 86.01 & 87.02 & {93.32} & \underline{85.77} & 96.57 & 64.41 & 82.20 & 72.36 & 88.73 & 64.93 & 82.28 & 74.50 & 77.69 & 69.37 & 84.09 & 72.26  \\
    &{PeFAD} & 88.77 & 94.74 & \textbf{97.22} & \textbf{91.34} & 97.93 & 97.46 & \textbf{98.35} & \textbf{97.68} & 87.71 & 89.78 & \textbf{94.43} & \textbf{88.73} & 73.42 & 87.31 & \textbf{92.61} & \textbf{78.94}\\

    \bottomrule
    \end{tabular}
  }
\end{table*}

\section{EXPERIMENTS}
\label{sec:EXPS}

\subsection{Datasets and Experiment Setup}
\subsubsection{Datasets} 
We conduct experiments on four real-world time series anomaly detection datasets: SMD, PSM, SWaT, and MSL. The 4 datasets are widely used by existing studies and are collected from various real-world domains, covering Internet data, server operational data, critical infrastructure system data, and spacecraft monitoring system events.
\begin{itemize}[leftmargin=12pt]
    \item \textbf{SMD.} Server Machine Dataset (SMD)~\cite{su2019robust} is a 5-week-long dataset collected from a large Internet company with 38 feature dimensions.
    \item \textbf{PSM.} Pooled Server Metrics (PSM) dataset~\cite{abdulaal2021practical} is collected from multiple application servers at eBay with 25 feature dimensions. 
    \item \textbf{SWaT.} Secure Water Treatment (SWaT) dataset~\cite{mathur2016swat} is obtained from 51 sensors of the critical infrastructure system under continuous operations.
    \item \textbf{MSL.} Mars Science Laboratory rover (MSL) dataset~\cite{hundman2018detecting} contains the telemetry anomaly data derived from the incident surprise anomaly reports of spacecraft monitoring systems with 55 feature dimensions. 
\end{itemize}

\subsubsection{Baselines}
We compare PeFAD with the following 12 baselines including classical methods: OCSVM~\cite{tax2004support}, Isolation Forest (IF)~\cite{liu2008isolation} LOF~\cite{breunig2000lof}, GANF~\cite{dai2022graph}, MTGFLOW~\cite{zhou2023detecting}, centralized reconstruction-based methods: Anomaly Transformer (AT)~\cite{xu2022anomaly}, TimesNet~\cite{wu2022timesnet}, and FPT~\cite{zhou2023one}, centralized prediction-based methods: Autoformer~\cite{wu2021autoformer}, Informer~\cite{zhou2021informer}, and FEDformer~\cite{zhou2022fedformer}. In addition, we transform centralized methods with FedAvg~\cite{mcmahan2017communication} into their federated version:  AT$_{fl}$, Autoformer$_{fl}$, Informer~\cite{zhou2021informer}, and FEDformer~\cite{zhou2022fedformer}, TimesNet$_{fl}$, and FPT$_{fl}$. We also compare PeFAD with the best performing model (i.e., DeepSVDD) in FedTADBench~\cite{liu2022fedtadbench}.


\subsubsection{Evaluation Metrics}
Precision (P), Recall (R), F1-Score (F1), and AUC-ROC (AUC, the Area Under the Receiver Operating Characteristic curve) are adopted as the evaluation metrics. A higher value of the metrics means a better performance. 


\subsubsection{Implementation Details.} 
We implement our model with the PyTorch framework on NVIDIA RTX 3090 GPU. The pre-trained language models (i.e., GPT2, BERT, ALBERT, RoBERTa, DeBERTa, DistillBERT, and Electra) are downloaded from Huggingface. 
We first split the time series into consecutive non-overlapping segments by sliding window~\cite{shen2020timeseries}. The patch length and batch size are set to 10 and 32, respectively. Adam is adopted for optimization. We adopt the widely-used point adjustment strategy~\cite{shen2020timeseries,su2019robust,xu2018unsupervised}. We employ GPT2 as the PLM, where the first eight layers of GPT2 are used for training.  $\lambda$ is set to $1e1$, $2e0$, $2e3$, and $15e4$ for SMD, PSM, SWaT, and MSL, respectively. The threshold $r$ for SMD, MSL, PSM, and SWaT is set to $0.5$, $2$, $1$, and $1$, respectively. 

\subsection{The Main Result}
Table~\ref{tab:main_res} shows the performance comparison among different methods under the federated and centralized settings on four datasets. 
In the federated setting, the best performance is marked in bold and the second-best result is underlined. In the centralized setting, the best performance is marked in red.
We use PeFAD$_c$ to represent the centralized version of PeFAD.  

From Table~\ref{tab:main_res}, one can see that PeFAD achieves the best performance in terms of F1-Score and AUC compared to all federated baselines on all four datasets, and even exceeds all centralized baselines on SMD and PSM datasets. More specifically, PeFAD outperforms the federated baselines by an average of \textbf{3.83\%--28.74\%} and \textbf{3.42\%--19.82\%} in terms of F1-Score and AUC metrics, respectively. 
Moreover, one can observe that PeFAD$_c$ shows the best overall performance under the centralized setup.
FPT exhibits sub-optimal integrated performance in the centralized baselines, which also utilizes PLM. It demonstrates the effectiveness of PLM in the task of time series anomaly detection. However, the performance of FPT under the federated setting shows a degradation. 
For example, PeFAD outperforms FPT$_{fl}$ by \textbf{9.85\%} and \textbf{7.37\%} for F1-Score and AUC metrics on SMD, respectively.
This might be attributed to the fact that FPT does not employ parameter-efficient tuning methods suitable for federated training, and the redundant parameters may affect the model performance.

A decreasing trend of performance is observed when transferring the baseline models from the centralized setting to federated setting, indicating that time series anomaly detection has become more difficult in federated environment. This is possibly due to the data sharing restrictions, which limit clients to use less data for model training. However, PeFAD demonstrates the best overall performance in both federated and centralized settings, indicating its robust adaptability to environmental changes. It can also be observed that in some cases (i.e. SMD dataset), the performance of PeFAD surpasses PeFAD$_c$. This may be attributed to the diversity of time series data. Through federated learning, models trained on each local device can better capture the diversity of its local data. Clients can obtain more adaptive thresholds based on the characteristics of their local data, whereas a single threshold obtained under the centralized setup may fail to accommodate the entire data.
\begin{figure}[t]
	\centering
	\begin{subfigure}{0.49\linewidth}
		\centering
		\includegraphics[width=1\linewidth]{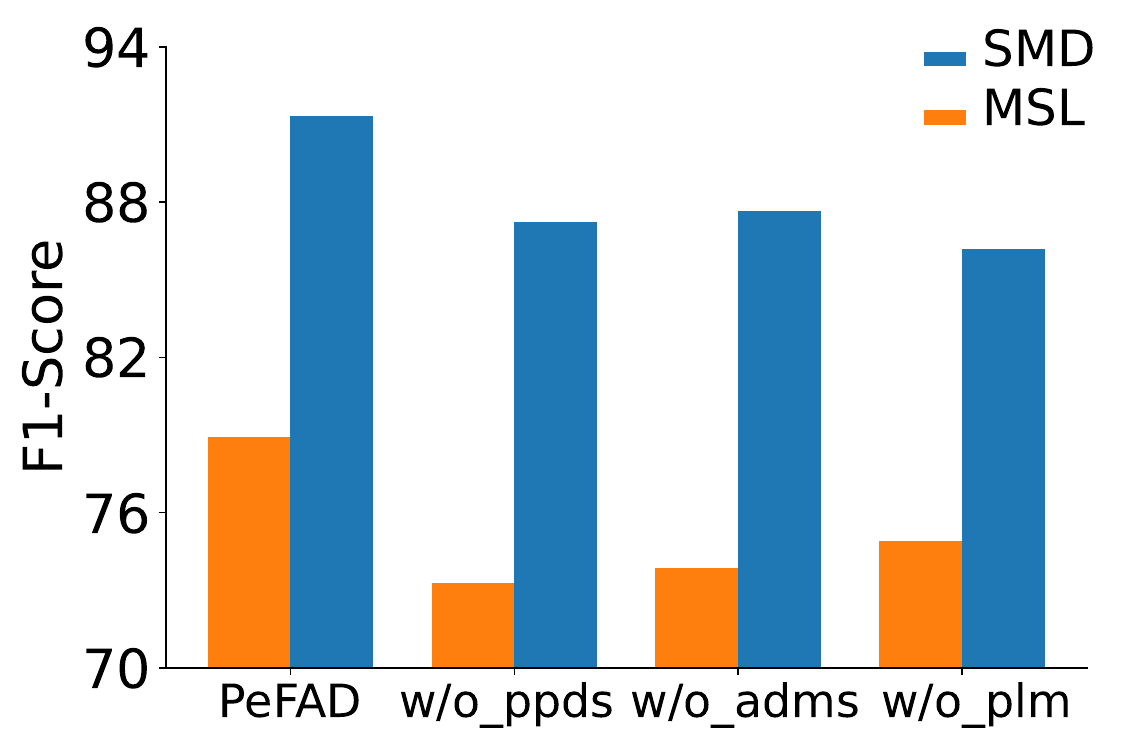}
    \captionsetup{font=small}
		\caption{F1-Score}
		\label{chutian1}
	\end{subfigure}
	\begin{subfigure}{0.49\linewidth}
		\centering
		\includegraphics[width=1\linewidth]{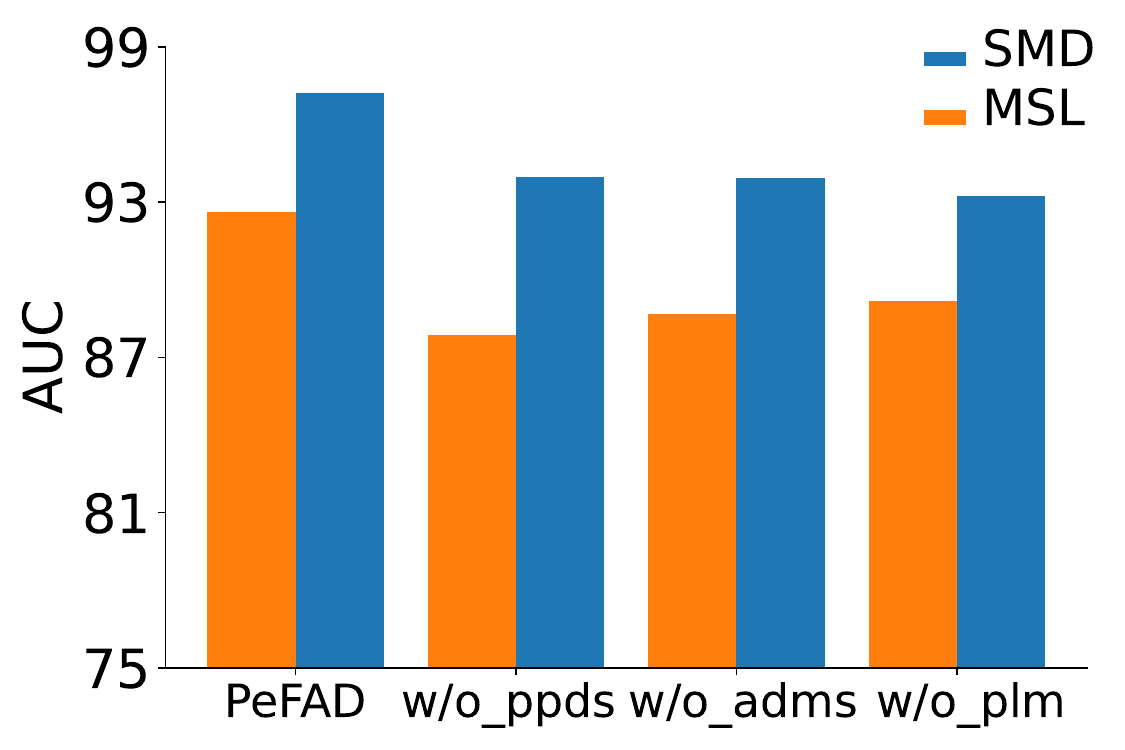}
  \captionsetup{font=small}
		\caption{AUC}
	\end{subfigure}
    \caption{Ablation study results of PeFAD and its variants}
    \label{fig:ablation}
    
\end{figure}

\subsection{Ablation Study}
To gain insight into the effects of key aspects of PeFAD, 
we compare the performance of PeFAD with its four variants as follows. $\ w/o\_ppds$: PeFAD without privacy-preserving shared dataset synthesis (PPDS) mechanism; $w/o\_adms$: PeFAD without anomaly-driven mask selection (ADMS) strategy, where ADMS is replaced with random masking; $w/o\_plm$: PeFAD without pre-trained language model (PLM) and it is replaced by transformer. 
We conduct experiments on SMD and MSL, which have the largest and smallest data volumes, respectively. 
The results are shown in Figure ~\ref{fig:ablation}. On both datasets, PeFAD always outperforms its counterparts without PPDS, ADMS, and PLM. It shows the three components are all useful for time series anomaly detection since removing any one of them will remarkably decrease the performance. 




\begin{table}[t]
\caption{Effect of various tuning strategies}
\centering
\renewcommand{\arraystretch}{1.13} 
\begin{tabular}{cccc|ccccc}
\toprule
\multirow{2}{*}{Methods} & \multicolumn{3}{c}{SMD} &\multicolumn{3}{c}{MSL} \\

\cmidrule(lr){2-7}

 & \small{AUC} & \small{F1} & \footnotesize{\makecell{Comm \\ Cost (GB)}} & \small{AUC} & \small{F1} & \footnotesize{\makecell{Comm \\ Cost (GB)}}\\
\midrule
FPT$_{fl}$ & 89.85 & 81.49 & 3.060 & 85.52 & 71.85 & 6.120\\
w/o\_ft & 94.74 & 88.18 & 0.000 & 90.47 & 76.17 & 0.000 \\
PeFAD\_t1l & 96.60 & 90.28 & 0.624 & \textbf{92.61} & \textbf{78.94} & 0.312 \\
PeFAD\_t2l & 96.88 & 90.76 & 1.216 &\underline{91.82} & \underline{77.96} & 0.608 \\
PeFAD\_t3l & \textbf{97.22}& \underline{91.34} & 1.800 & 91.62 &77.64& 0.900 \\
PeFAD\_t4l & \underline{97.16} & \textbf{91.37} & 2.384 & 90.10 & 76.30& 1.192 \\
PeFAD\_t5l & 96.93 & 90.80 & 2.976 & 89.63 & 75.70 & 1.488 \\
PeFAD\_t6l & 97.01 & 90.79 & 3.560 & 88.74 & 74.26 & 1.780 \\
PeFAD\_t7l & 97.00 & 90.74 & 4.144 & 87.93 & 75.32& 2.072 \\
PeFAD\_fft & 97.07 & 90.91 & 6.648 & 87.06 & 72.38 & 3.324 \\
\bottomrule
\label{tab:tuning}
\end{tabular}
\end{table}

\subsection{Effect of Tuning Strategies and PLMs}
\subsubsection{Effect of various tuning strategies} 
To test the effect of different tuning strategies of PLM, we compare PeFAD with strategies of fine-tuning different numbers of PLM layers, including no fine-tuning (w/o\_ft), tuning the last one to seven layers of PLM (PeFAD\_t1l - PeFAD\_t7l), and fully fine-tuning (PeFAD\_fft). The result is shown in Table~\ref{tab:tuning}. We use GPT2-based FPT$_{fl}$ as a reference.
One can observe that freezing the first layers while fine-tuning the last few layers is a reasonable tuning strategy. By freezing the first layers, the model retains the ability to understand generalized knowledge, and fine-tuning the last few layers facilitates the model's adaptation to downstream tasks, enabling the transfer of domain-specific knowledge from the pre-trained model to the time series anomaly detection task. 
Specifically, for the SMD dataset with more training data, PeFAD remains relatively stable with different tuning layers, and achieves optimal performance when tuning the last 3 and 4 layers. For the smaller MSL dataset, the model performance decreases with the increase of tuning layers, reaching optimal performance when tuning the last layer.
The experiments on other datasets are provided in the appendix due to space limitation. In PeFAD, we choose to fine-tune the last layer for MSL and fine-tune the last three layers for the other datasets.

The result shows that our approach consistently outperforms FPT regardless of the number of tuning layers. Compared with FPT, PeFAD achieves the performance improvement of \textbf{9.85\%} and \textbf{7.09\%} in terms of F1-Score on SMD and MSL, respectively. PeFAD reduces the communication cost by \textbf{41.2\%} and \textbf{94.9\%}, which shows the efficiency of PeFAD and the effectiveness of the proposed parameter-efficient federated training module. Furthermore, PeFAD without fine-tuning (w/o\_ft) outperforms all federated baselines on both datasets, which demonstrates the superior cross-modality knowledge transfer ability of PLM. 
PeFAD\_fft does not achieve the best performance on both datasets while tuning less, especially last few layers, works better. This is because the initial layers of PLM contain generic knowledge and the last layers are better suited to learn task-specific information. However, due to the scarcity of anomalous data, fully fine-tuning may increase the risk of overfitting, leading to performance degradation.


        

\subsubsection{Effect of various PLMs} 
\begin{figure}[!t]
	\centering
	 	\begin{subfigure}{0.49\linewidth}
		\includegraphics[width=1\linewidth]{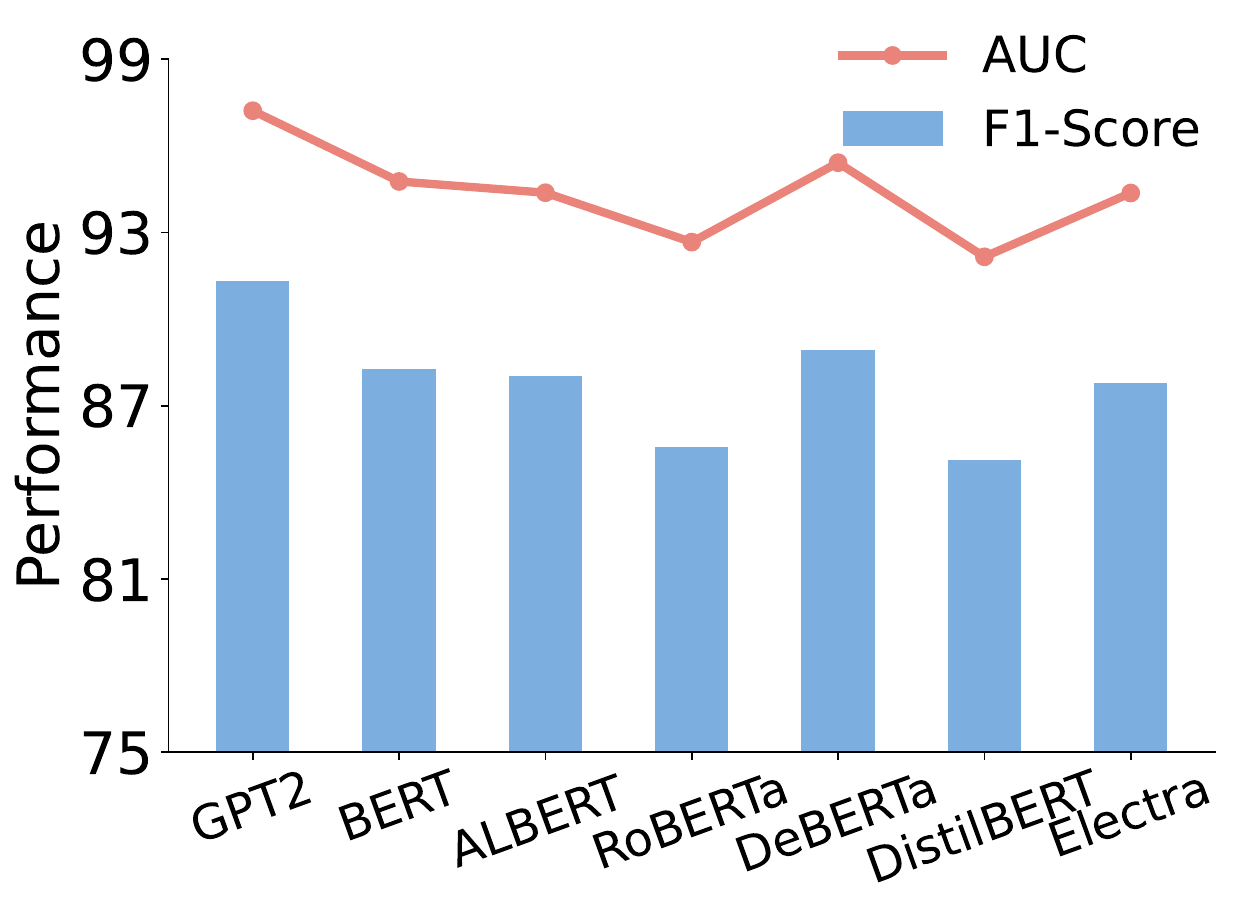}
  \captionsetup{font=small}
		\caption{SMD}
	\end{subfigure}
	\begin{subfigure}{0.49\linewidth}
		\centering
		\includegraphics[width=1\linewidth]{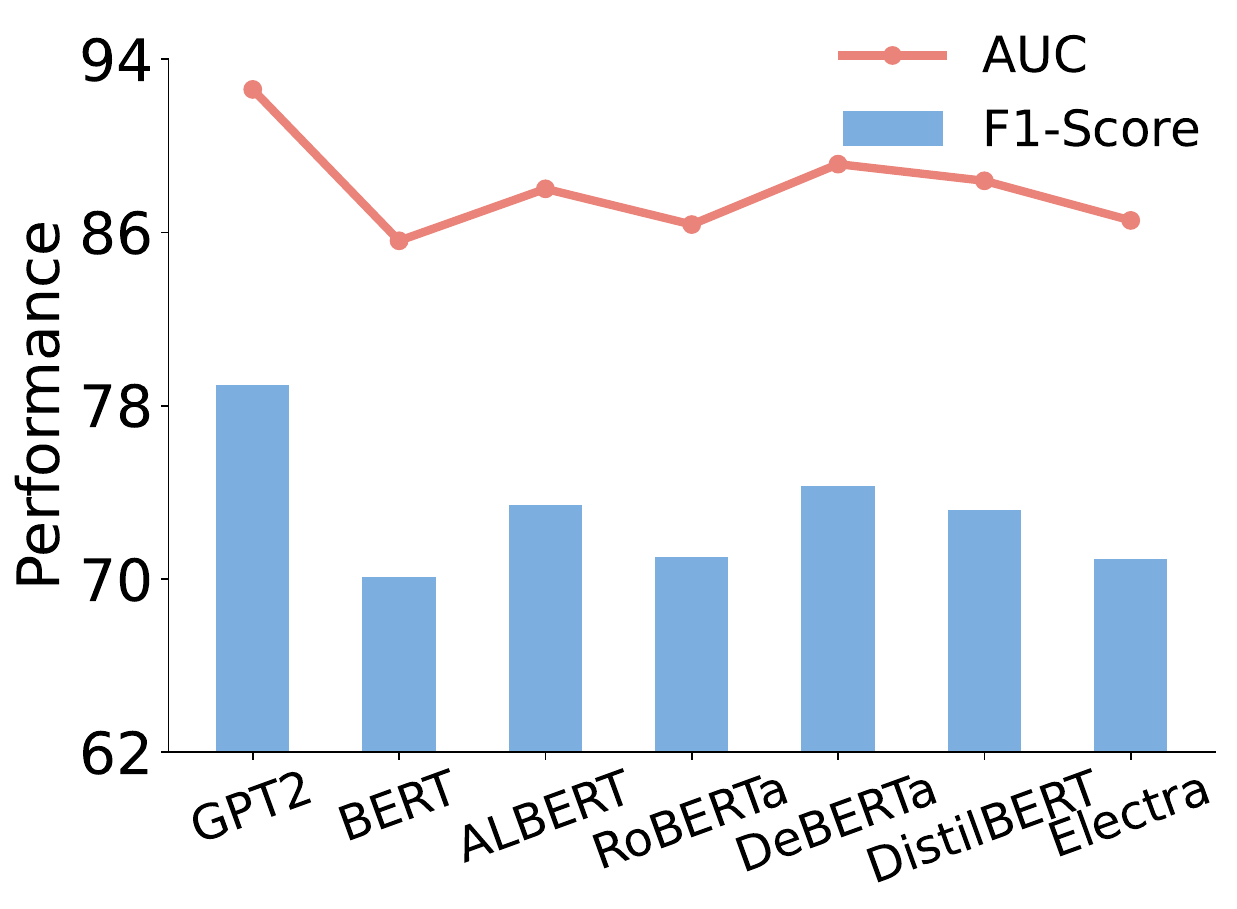}
  \captionsetup{font=small}
		\caption{MSL}
	\end{subfigure}
 \caption{Effect of various PLMs on model performance}
\label{fig:plm_models}
\end{figure}

Next, we study the effect of using different PLMs on the model performance. We compare seven mainstream pre-trained models, i.e., BERT, ALBERT, RoBERTa, DeBERTa, DistilBERT, and Electra. The results are presented in Figure~\ref{fig:plm_models}. One can see that GPT2 achieves the best performance followed by DeBERTa. Compared to other PLMs, GPT2 improves the performance by up to \textbf{6.22\% }and \textbf{5.06\%} on F1-Score and AUC metrics on SMD, respectively. On the MSL dataset, the F1-Score and AUC values are improved by up to \textbf{8.84\%} and \textbf{6.99\%}, respectively. This is because GPT2 has been exposed to a broader range of contexts during pre-training, enabling it to learn from time series more effectively.

\subsection{Parameter Sensitivity Analysis}
\subsubsection{Effect of various mask ratio $r_m$ and patch length $l_p$} 
\begin{figure}[h]
	\centering
	\begin{subfigure}[b]{0.475\linewidth}
		\includegraphics[width=1\linewidth]{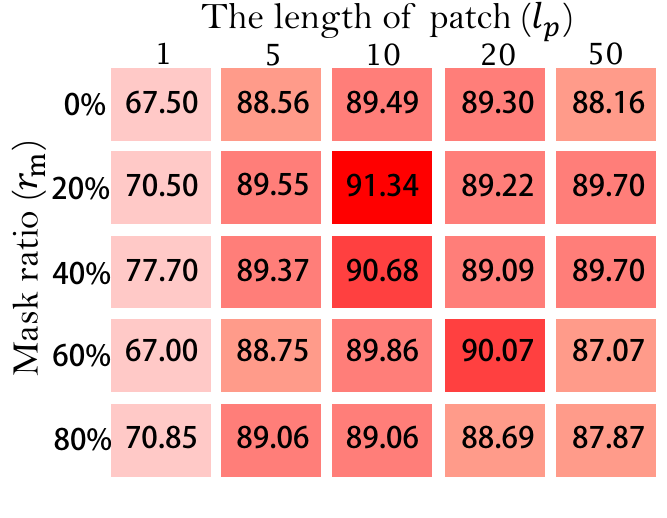}
            \captionsetup{font=small}
		\caption{Effect of mask and patch}
            \label{fig:mask_patch}
	\end{subfigure}\ \ \ 
    \begin{subfigure}[b]{0.49\linewidth}
		\includegraphics[width=1\linewidth]{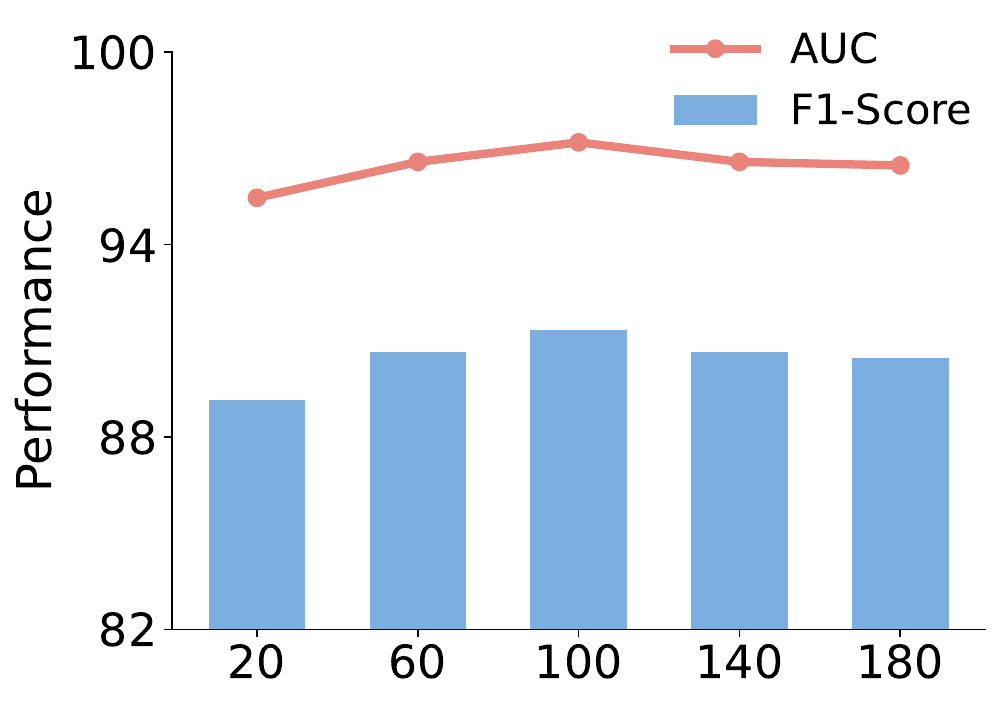}
            \captionsetup{font=small}
		\caption{Effect of synthetic data length}
            \label{fig:shared_data}
	\end{subfigure}

 \caption{Parameter sensitivity analysis on SMD dataset}  
 \label{fig:param_analyse}
\end{figure}

We next study the sensitivity of the model to the mask ratio $r_m$ and patch length $l_p$, We only give the result of F1-Score on SMD as an example due to space limitation, as shown in Figure~\ref{fig:param_analyse}(\subref{fig:mask_patch}). 
One can observe that the incorporation of masking or patching mechanisms can improve the model performance, demonstrating the effectiveness of these two mechanisms. 
As the $r_m$ and $l_p$ increase, the model performance first improves and then declines. The optimal model performance is achieved when $r_m$ is 20\% and $l_p$ is 10.

\subsubsection{Effect of synthetic series length} 
We next investigate the effect of synthetic data length on model performance, and the result is shown in Figure~\ref{fig:param_analyse}(\subref{fig:shared_data}). Specifically, we vary the length of the synthetic time series for each client on the SMD dataset.
We observe that the F1-Score curve first increases and then drops slightly. Generally, the result demonstrates that the model obtains the best performance when the length of the synthetic time series is set to 100. With the increase of length from 20 to 100, the synthetic time series may bring more useful information, which facilitates the model with more effective representation learning. 
However, a too large length value will lead to performance decline. This is because longer synthetic time series may bring redundant or noisy information, which degrades the model performance.

\subsection{Case Study}
\begin{figure}
	\begin{subfigure}{0.495\linewidth}
		\includegraphics[width=0.97\linewidth]{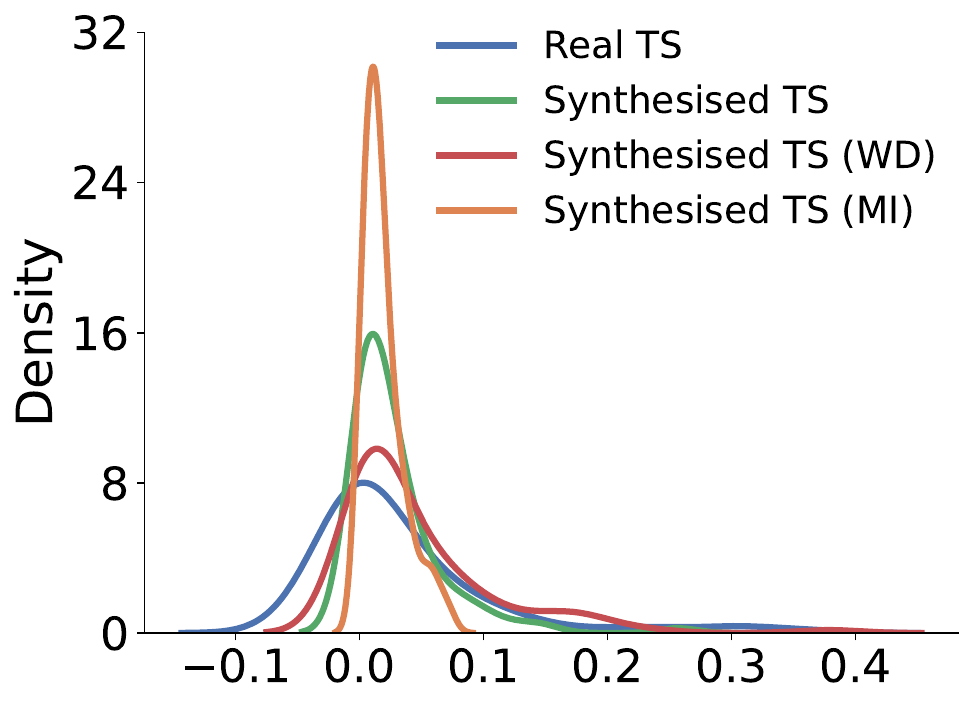}
            \captionsetup{font=scriptsize}
		\caption{The KDE of real and synthesised TS}
        \label{fig:KDE}
	\end{subfigure}
	\begin{subfigure}{0.495\linewidth}
		\centering
		\includegraphics[width=1.03\linewidth]{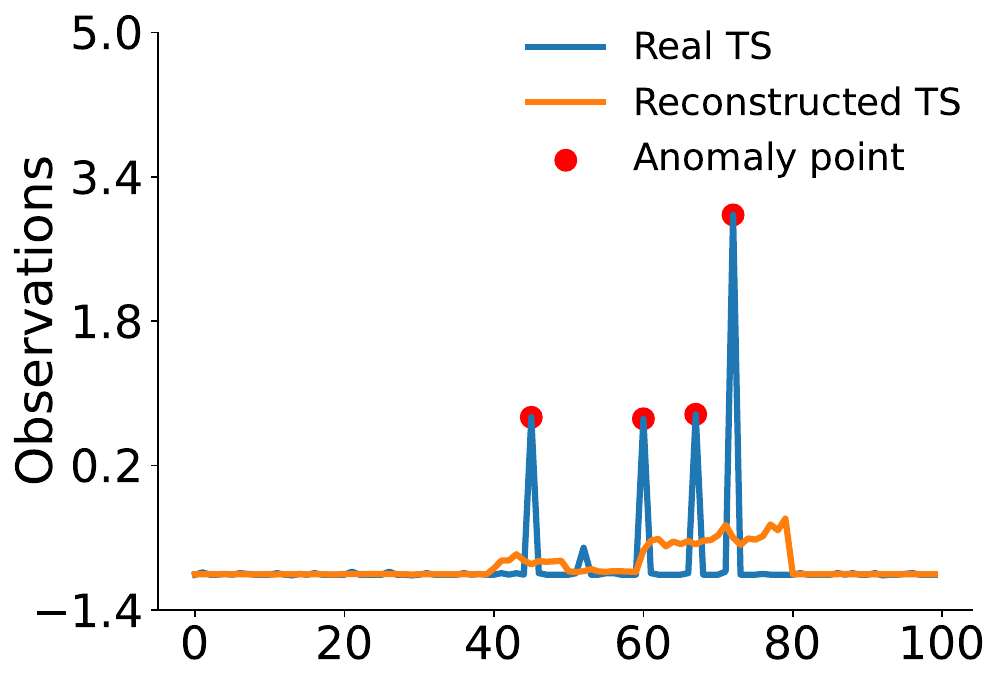}
              \captionsetup{font=scriptsize}
		\caption{An reconstruction example in testing}
        \label{fig:test_recon2}
	\end{subfigure} 
	\caption{The example of data synthesis, time series reconstruction and anomaly detection within the client from SMD dataset. }
	\label{fig:case_study}
\end{figure}

To intuitively show the effectiveness of the proposed PeFAD, we provide a case study on SMD, as illustrated in Figure~\ref{fig:case_study}. Figure~\ref{fig:case_study}(\subref{fig:KDE}) shows the distribution of the real and synthesized time series, estimated by Kernel Density Estimation.
The blue curve in the figure represents the real time series, the orange curve represents the synthesized time series obtained solely through mutual information (MI) constraint, the red curve represents the synthesized time series obtained solely through Wasserstein distance (WD) constraint, and the green curve represents the time series synthesized under the combined constraints of MI and WD.
One can see that the orange curve exhibits a significant difference from the blue curve, while the red curve closely resemble the real distribution (blue curve). This is because solely reducing mutual information neglects considerations on the quality of the synthesized data.
However, the green curve both ensures distributional similarity and protects the privacy of the data through mutual information. 

Figure~\ref{fig:case_study}(\subref{fig:test_recon2}) shows an example of time series reconstruction and anomaly detection on the SMD dataset during testing within the client. One can observe that the estimated values at normal points closely approximate the true values, while at anomalous points, the estimates align more closely with reasonable values unaffected by anomalies. Thus the anomalies in the time series are successfully identified by assessing the disparity between estimated and actual values. This is probably attributed to the proposed ADMS strategy and the PPDS mechanism, which empower the model to better adapting to complex patterns, thereby contributing to the effectiveness of time series anomaly detection.

\section{DISCUSSION}\label{sec:DISSCUSS}
We conduct comprehensive experiments, showing that PeFAD outperforms state-of-the-art baselines in terms of both centralized and federated methods. The results demonstrate the powerful representation learning capability of PLM. In addition, the proposed PPDS module also improves stability under FL. The ablation study further verifies the effectiveness of the three major components of PeFAD (i.e., PLM, ADMS, and PPDS). 
Specifically, the ADMS strategy makes the model focus more on changing regions in the time series by capturing intra- and inter-patch dynamics changes. As time series often change frequently with time evolving, enhancing the model's capability in learning such changes can facilitate the proposed model to learn representative features. Moreover, the PPDS mechanism helps the model achieve more consistent client updates, thereby improving the performance and stability of the aggregated global model.
Moreover, we also verify that the proposed efficient tuning strategy reduces communication overhead effectively.

\section{CONCLUSION}\label{sec:CONCLU}

This work presents PeFAD, a federated learning framework for time series anomaly detection. Different from previous methods, we aim to leverage the generic knowledge and the contextual understanding capability of the pre-trained language model to address the data scarcity problem. To alleviate the communication and computation burden in federated learning brought by PLM, we propose a parameter-efficient federated training module, where clients only need to fine-tune and transmit small-scale parameters. Moreover, PeFAD features a novel anomaly-driven mask selection strategy to refine the quality of time series reconstruction, thereby improving the robustness of anomaly detection. In order to address the issue of client heterogeneity, a privacy-preserving shared dataset synthesis mechanism is also proposed, enabling clients to learn more consistent and comprehensive information.
Extensive experiments on four real work datasets show the effectiveness and efficiency of the proposed PeFAD. 



\bibliographystyle{ACM-Reference-Format}
\bibliography{sample-base}
\appendix
\appendix
\section{Appendix}

\subsection{Evaluation Metrics}
We adopt Precision, F1-Score, Recall, and AUC-ROC (AUC) as the evaluation metrics, which are defined as follows.
\begin{equation}
\begin{split}
    &Precision = \frac{TP}{TP + FP}, \\ 
    &F1{-}Score = \frac{2 \times precision \times recall}{precision+ recall},\\
    &Recall = \frac{TP}{TP + FN}, \\
    &AUC = \int_{0}^{1} ROC_{curve}\ \, d\ FPR,
\end{split}
\end{equation}
where TP represents True Positive, FP denotes False Positive, and FN is False Negative. FPR (False Positive Rate) represents the proportion of negative instances that are incorrectly classified as positive. 
AUC represents the Area Under the Receiver Operating Characteristic (ROC) curve. 


\subsection{Additional Experiments}
\ 

\subsubsection{Ablation Study.} 
\begin{figure}[h]
\begin{subfigure}{0.495\linewidth}
    \centering
    \includegraphics[width=1\linewidth]{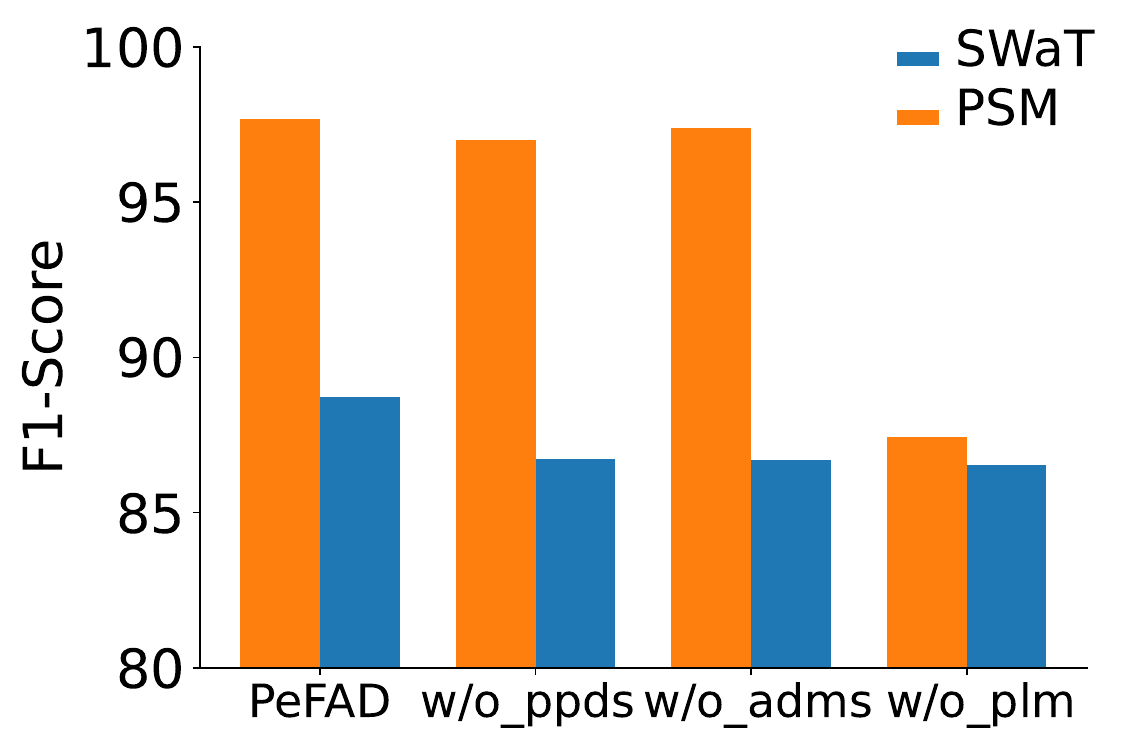}
    \caption{F1-Score}  
\end{subfigure}
    \begin{subfigure}{0.495\linewidth}
\includegraphics[width=1\linewidth]{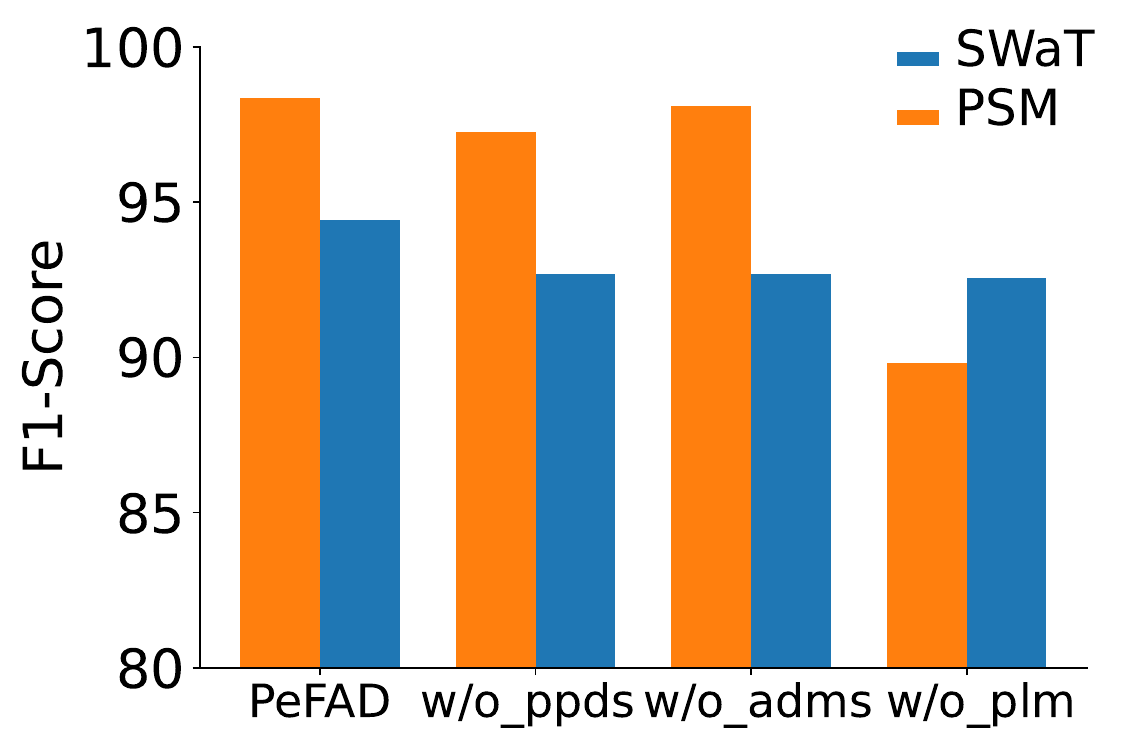}
\caption{AUC}
    \end{subfigure}
    \caption{Ablation study results of PeFAD and its variants.}
    \label{fig:ablation_swat_psm}   
\end{figure}

The results of the ablation experiments on the SWaT dataset and PSM dataset are shown in Figure ~\ref{fig:ablation_swat_psm}. The results show that PeFAD outperforms the other 3 ablation variants in both AUC and F1-Score metrics. The variant without PLM performs the worst, which demonstrates the effectiveness of PLM on the task of federated anomaly detection.

To further explore the effects of various variants on PeFAD performance, we conducted more detailed ablation experiments.
\begin{itemize}[leftmargin=12pt]
    \item \textbf{$ w/o\_ppds$.} PeFAD without the shared dataset synthesis scheme.  
    \item \textbf{$ w/o\_adms$.} PeFAD without ADMS strategy replaced by random masking.
    \item \textbf{$ w/o\_plm.$} PeFAD without pre-train language model (PLM) replaced by transformer.
    \item \textbf{$ w/o\_(adms{-}intra)$.} PeFAD without intra-patch time series decomposition when calculating the anomaly score of patches, which means the hyper-parameter $\beta$ is equal to 0. 
    \item \textbf{$ w/o\_(adms{-}inter)$.} PeFAD without inter-patch similarity assessment when calculating the anomaly score of patches, which means the hyper-parameter $\beta$ is equal to 1. 
    \item \textbf{$w/o\_ppds\&adms$.} PeFAD without PPDS and ADMS. 

\end{itemize}

\begin{figure}
	\begin{subfigure}{0.495\linewidth}
		\centering
		\includegraphics[width=1\linewidth]{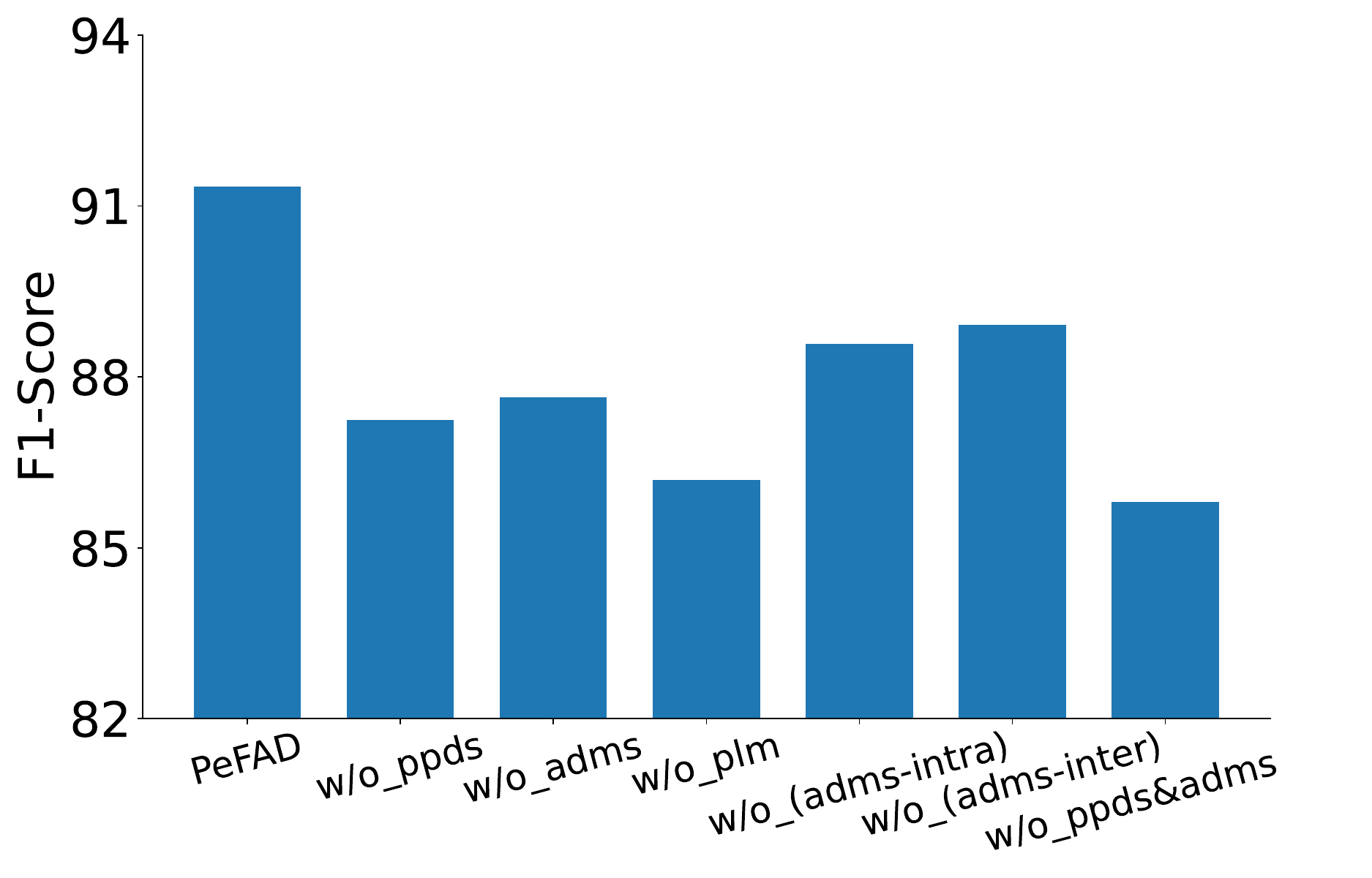}
		\caption{SMD dataset (F1-Score)}
	\end{subfigure}
	\begin{subfigure}{0.495\linewidth}
		\centering
		\includegraphics[width=1\linewidth]{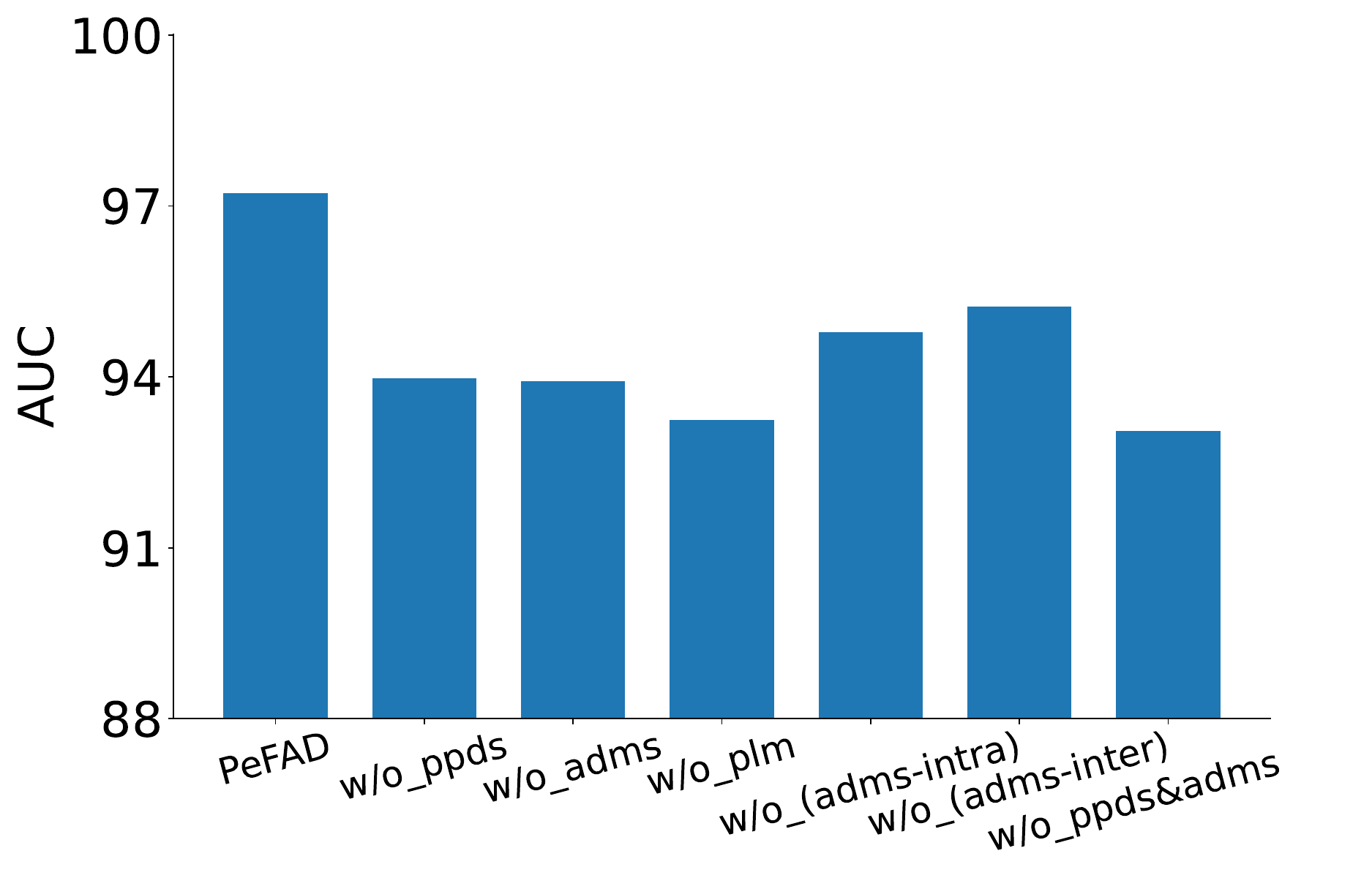}
		\caption{SMD dataset (AUC)}
  \label{fig:Precision_appendix}
	\end{subfigure}
	\begin{subfigure}{0.495\linewidth}
		\centering
		\includegraphics[width=1\linewidth]{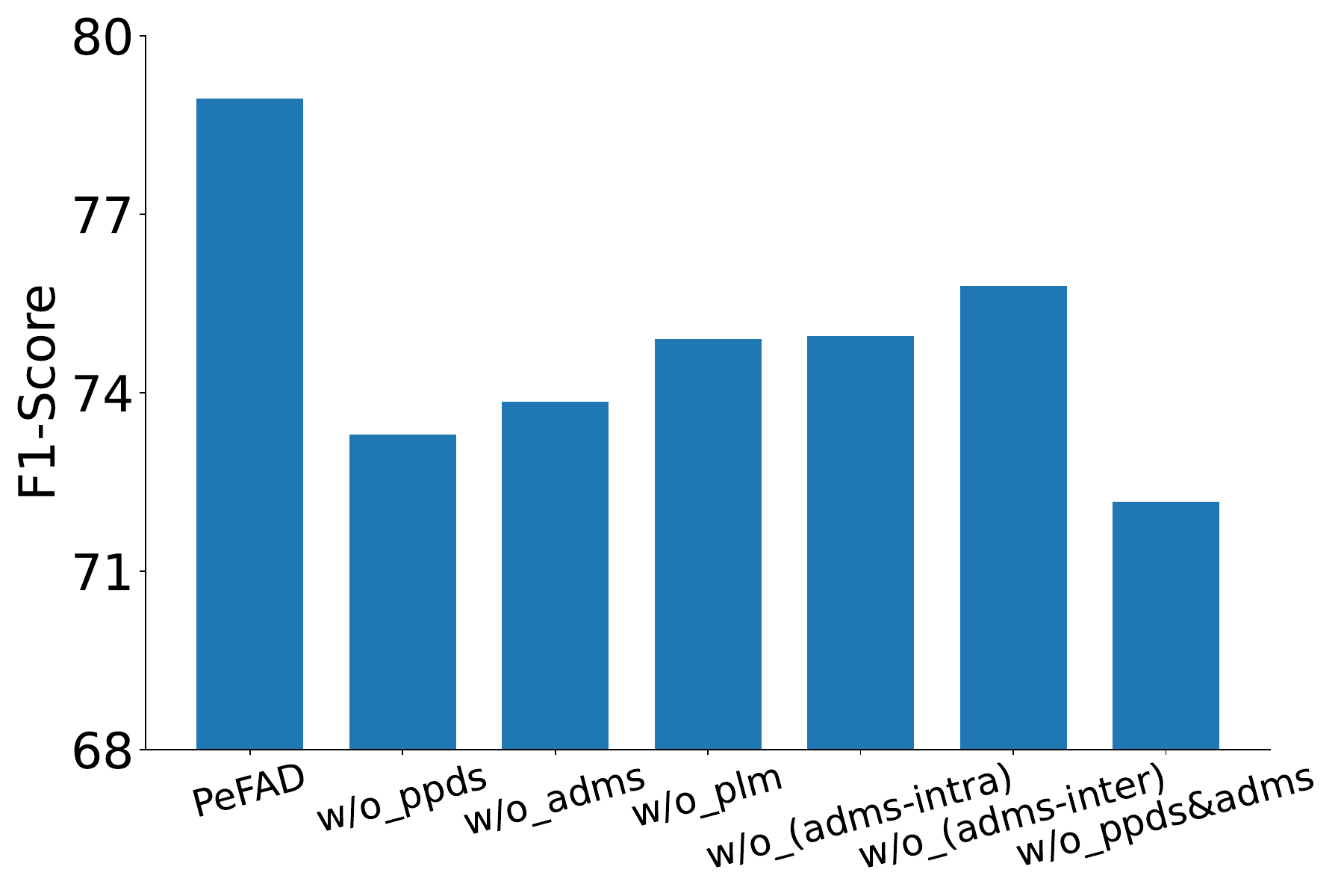}
		\caption{MSL dataset (F1-Score)}
	\end{subfigure}
	\begin{subfigure}{0.495\linewidth}
		\centering
		\includegraphics[width=1\linewidth]{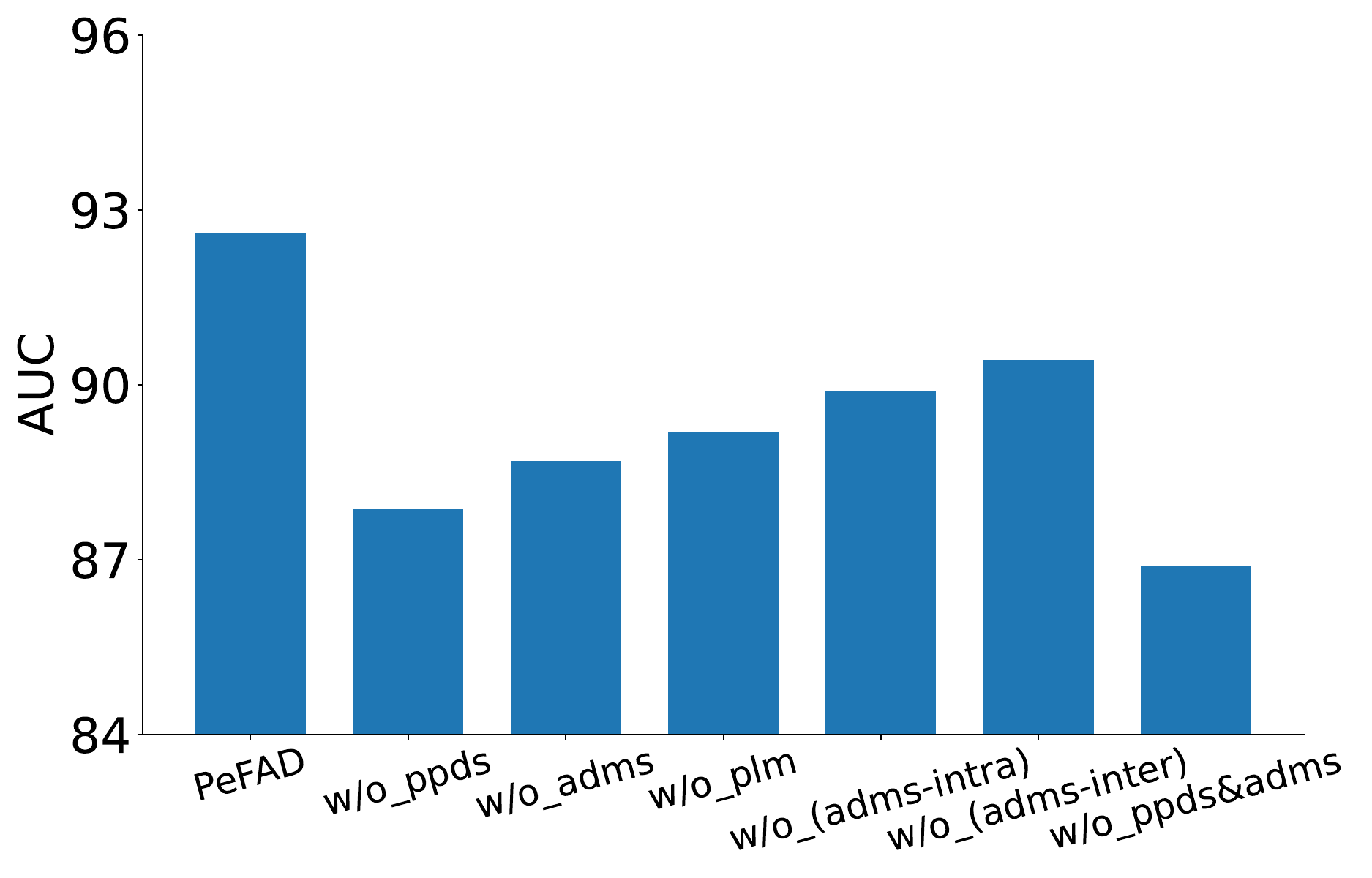}
		\caption{MSL dataset (AUC)}
  \label{fig:Precision_appendix}
	\end{subfigure}
	\caption{The ablation study results on SMD and MSL dataset}
	\label{fig:ablation_appendix}
\end{figure}

The results on the SMD and MSL datasets are shown in Figure~\ref{fig:ablation_appendix}. One can see that these four components all improve the anomaly detection performance of PeFAD. For example, removing these components decreases the F1-Score and AUC values by up to \textbf{6.77\%} and \textbf{5.72\%} on MSL, respectively. On both datasets, $w/o\_ppds\&adms$ performs the worst among all variants on both datasets, showing the benefit of PPDS mechanism and ADMS strategy. Further, $\ w/o\_plm$ performs second-worst in terms of F1-Score, indicating the validity of the PLM. Specifically, on both datasets, $\ w/o\_kd\&adms$ performs the worst among all variants. PeFAD outperforms $\ w/o\_kd\&adms$, improving the performance by up to $6.15\%$ and $4.95\%$ in terms of F1-Score and AUC, respectively

\subsubsection{Effect of Various Tuning Strategies.} 
We further investigate the effect of various tuning strategies on PSM and SWaT datasets. The results are shown in Table~\ref{tab:appendix_tuning}. 
It can be seen that the best choice for the PSM dataset is to fine-tune the last 3 layers, and for the SWaT dataset fully fine-tuning and fine-tuning the last three layers achieve similar performance. To reduce computation cost, we fine-tune the last three layers in PeFAD in practice for SWaT. In addition, compared to the FPT$_{fl}$, PeFAD which fine-tunes the last three layers shows better performance and lower communication overhead on both PSM and SWaT datasets, which demonstrates the effectiveness of the parameter-efficient federated training module.

\subsubsection{Effect of Different Fine-tuning Parameters.} 
\begin{figure}
\begin{subfigure}{0.495\linewidth}
    \centering
    \includegraphics[width=1\linewidth]{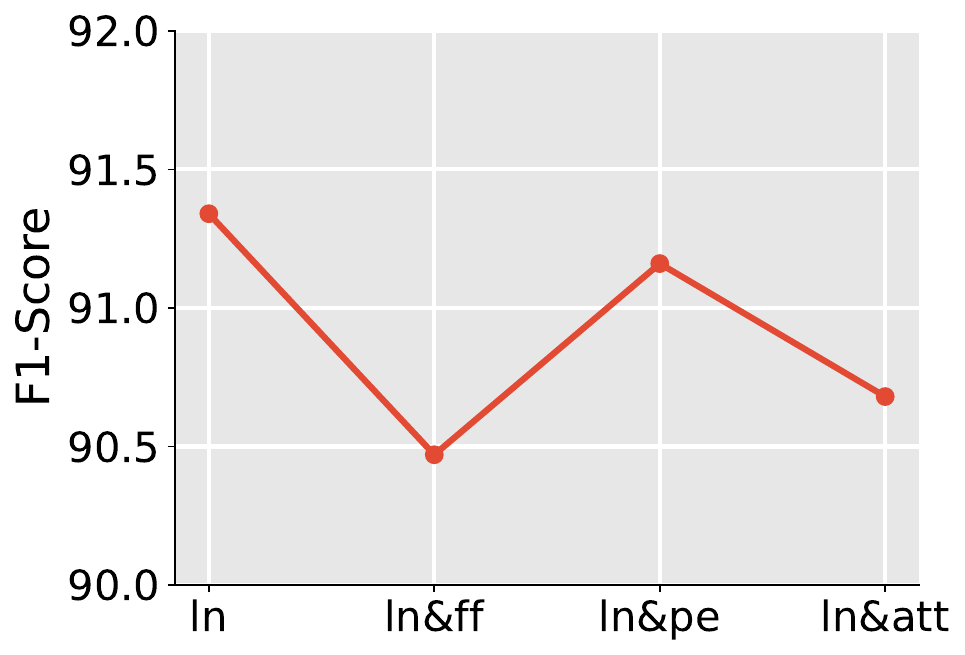}
    \caption{F1-Score}  
\end{subfigure}
    \begin{subfigure}{0.495\linewidth}
\includegraphics[width=1\linewidth]{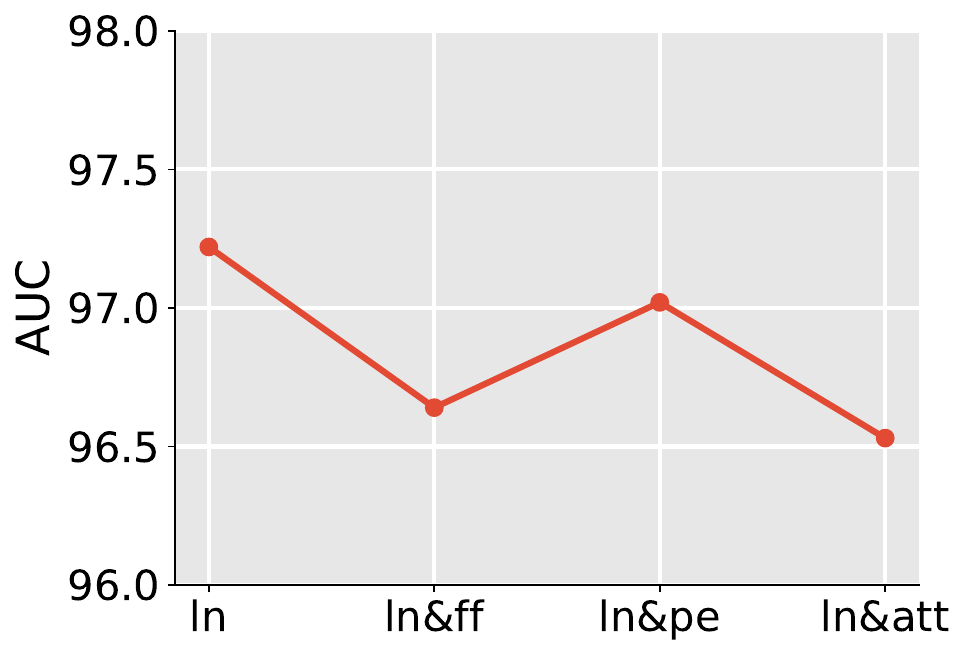}
\caption{AUC}
    \end{subfigure}
    \caption{The effect of different fine-tuning parameters}
    \label{fig:tuning_params}   
\end{figure}

We next study the effect of different fine-tuning parameters to assess the importance of different parameters in various layers. 
GPT2 consists of the following layers: the position embedding layer (pe), the layer norm (ln), the attention layer (att), and the feedforward layer (ff). We conduct experiments on the SMD dataset, and the result is shown in Fig~\ref{fig:tuning_params}. We only fine-tune the last three layers, and it can be observed that fine-tuning the blocks of pe, att, and ff is the optimal fine-tuning solution. It is because these blocks contain task-specific information and adjusting them allows the model to adapt to the nuances of the target domain or task.
\begin{table}[t]
\caption{Effect of various tuning strategies}
\renewcommand{\arraystretch}{1} 
\centering
\small
\begin{tabular}{cccc|ccccc}
\toprule
\multirow{2}{*}{Methods} & \multicolumn{3}{c}{PSM} &\multicolumn{3}{c}{SWaT} \\

\cmidrule(lr){2-7}

 & \small{AUC} & \small{F1} & \footnotesize{\makecell{Comm \\ Cost (GB)}} & \small{AUC} & \small{F1} & \footnotesize{\makecell{Comm \\ Cost (GB)}}\\
\midrule
FPT$_{fl}$ & 95.66 & 94.92 & 6.120 & 92.28 & 86.74 & 6.120\\
w/o\_ft & 97.02 & 96.31 & 0.000 & 91.33 & 84.97 & 0.000 \\
PeFAD\_t1l & 98.05 & 97.36 & 0.780 & {92.54} & {86.54} & 0.156 \\
PeFAD\_t2l & 98.08 & 97.46 & 1.520 &{94.15} & {88.53} & 0.304 \\
PeFAD\_t3l & \textbf{98.35}& \textbf{97.68} & 2.250 & \textbf{94.43} &\underline{88.73}& 0.450 \\
PeFAD\_t4l & {98.15} & {97.49} & 2.980& 94.20 & 88.63& 0.596 \\
PeFAD\_t5l & 98.23 & \underline{97.55} & 3.720 & 94.05 & 88.39 & 0.744 \\
PeFAD\_t6l & \underline{98.26} & 97.52 & 4.450 & 94.23 & 88.63 & 0.89 \\
PeFAD\_t7l & 98.16 & 97.39 & 5.180 & 94.19 & 88.56& 1.036 \\
PeFAD\_fft & 98.07 & 97.23 & 8.310 & \underline{94.29} & \textbf{88.75} & 1.662 \\
\bottomrule
\label{tab:appendix_tuning}
\end{tabular}
\end{table}

\begin{figure}
	\begin{subfigure}{0.495\linewidth}
		\centering
		\includegraphics[width=1\linewidth]{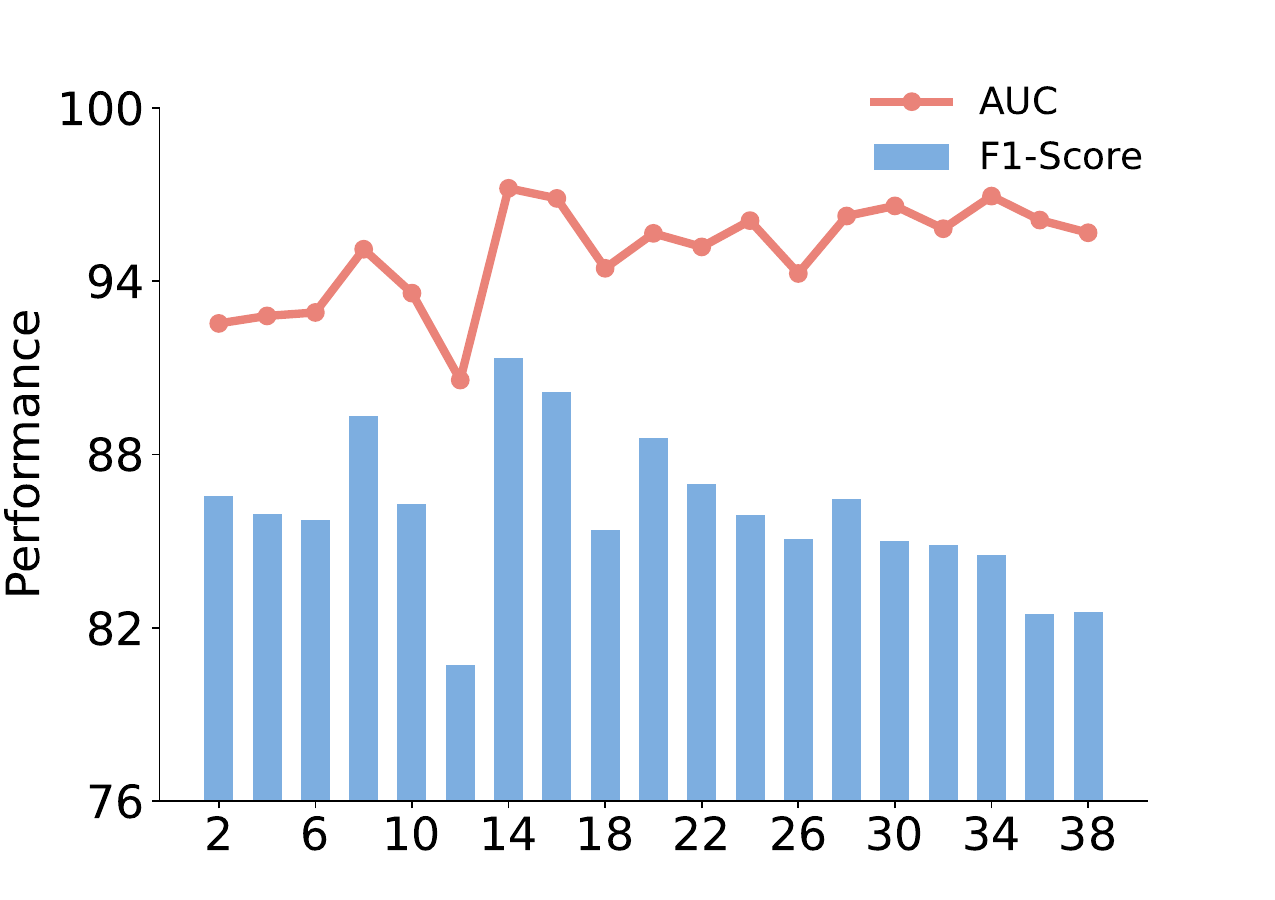}
  \captionsetup{font=scriptsize}
              
		\caption{The effect of client numbers}
        \label{fig:appendix_clientnums}
	\end{subfigure}
	\begin{subfigure}{0.495\linewidth}
		\centering
		\includegraphics[width=1\linewidth]{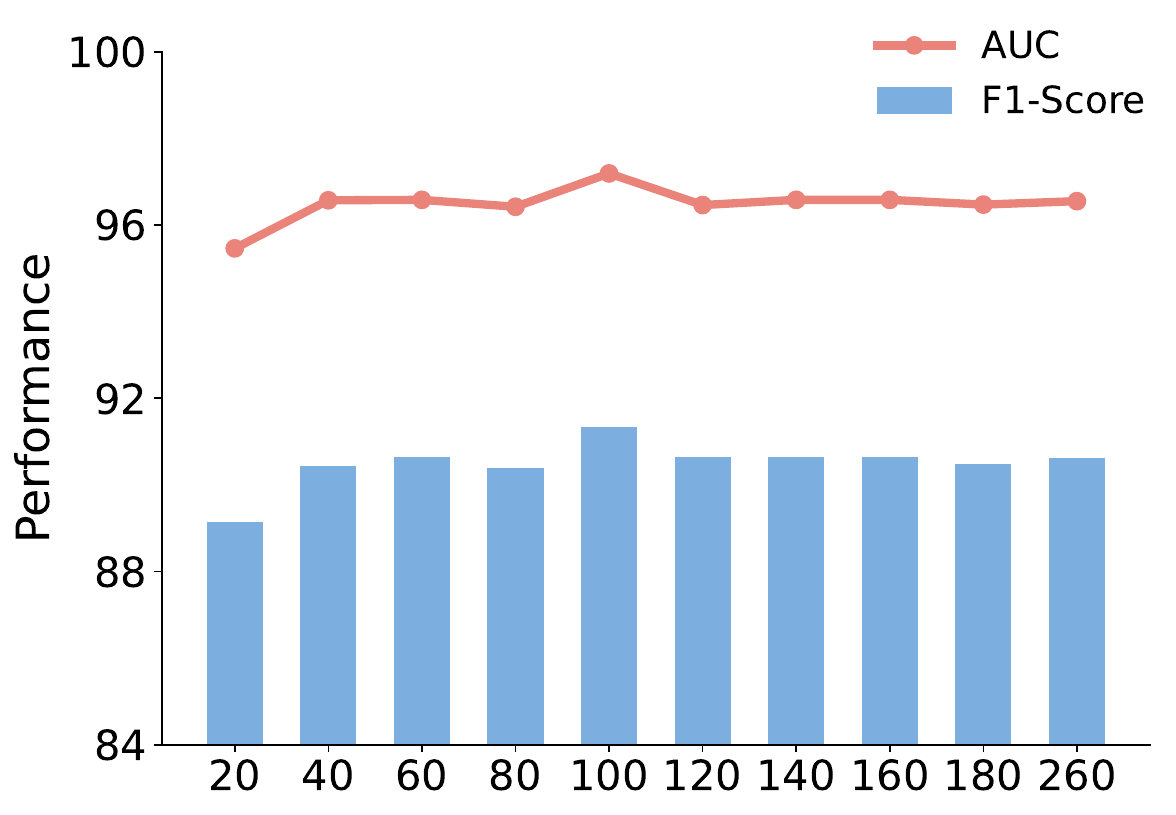}
  \captionsetup{font=scriptsize}
		\caption{The effect of synthetic data length}
        \label{fig:appendix_shared}
	\end{subfigure}
	\caption{Parameter sensitivity analysis}
	\label{fig:appendix_params}
\end{figure}

\subsubsection{Parameter Sensitivity Analysis.} 
\ 

\textbf{(1) Effect of client numbers.} We investigate the effect of client numbers on the model performance over SMD, the result is shown in Figure~\ref{fig:appendix_params}(\subref{fig:appendix_clientnums}).
We observe that the model achieves optimal performance when the number of clients is set to 14, and when the number of clients exceeds 14, the model performance decreases as the number of clients increases. This is because as the number of clients increases, the model may become more prone to overfitting each individual client. This could lead to an overall performance decline.

\textbf{(2) Effect of synthetic data length.}
We investigate the synthetic data length on model performance by varying the length of the client-synthesis time series on the SMD, the result is shown in Figure~\ref{fig:appendix_params}(\subref{fig:appendix_shared}).
One can observe that the model is relatively robust to the different sizes of the synthesized time series, and the model performs best when the length of synthesized time series is set to 100.

\begin{figure}
	\begin{subfigure}{0.495\linewidth}
            \includegraphics[width=1\linewidth]{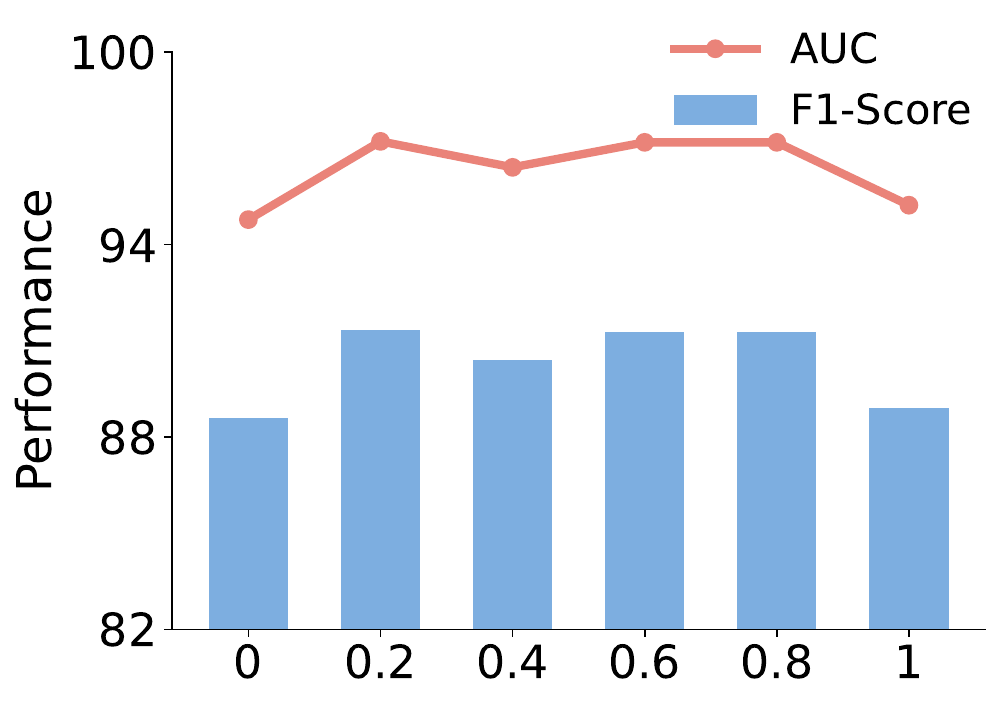}
            \captionsetup{font=scriptsize}
		\caption{Effect of $\beta$ in ADMS}
        \label{fig:train_recon}
	\end{subfigure}
	\begin{subfigure}{0.495\linewidth}
		\centering
		\includegraphics[width=1\linewidth]{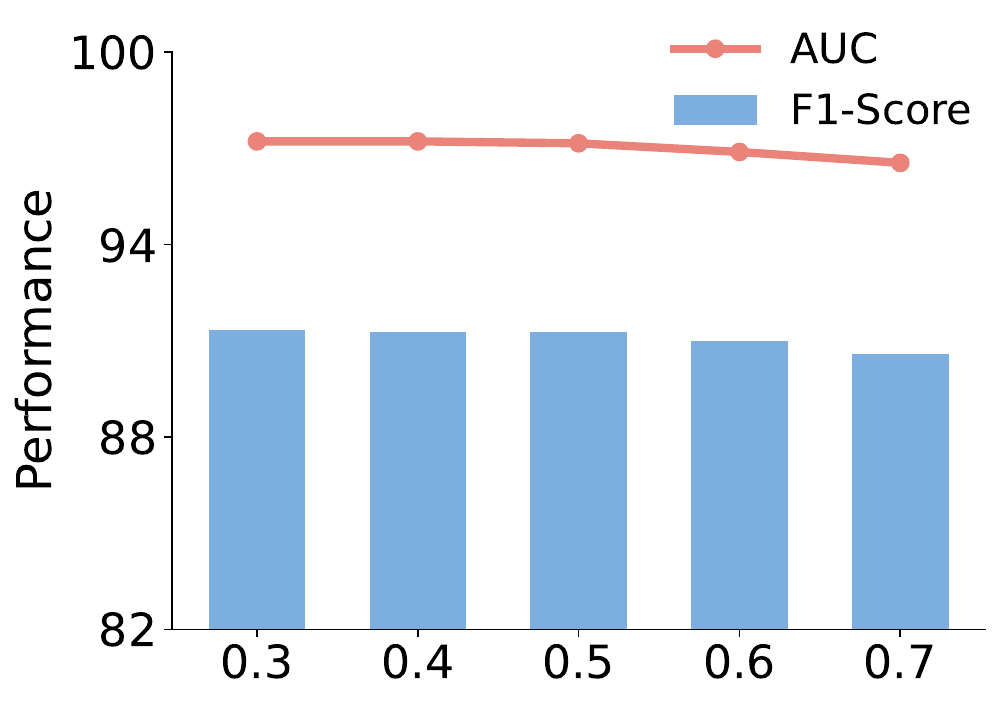}
              \captionsetup{font=scriptsize}
		\caption{Effect of $\alpha_2$ in PPDS}
        \label{fig:test_recon1}
	\end{subfigure}
	\caption{Effects of hyperparams in ADMS and PPDS. }
	\label{fig:appendix_hyper}
\end{figure}

\textbf{(3) Effect of hyperparameters in ADMS and PPDS.}
We conduct experiments on the hyperparameter (i.e., $\beta$ and $\alpha_2$) sensitivity of ADMS and PPDS on SMD, as shown in Figure~\ref{fig:appendix_hyper}. The results show that the fluctuation of the model's performance is not significant as the hyperparameters are varied, especially for the hyperparameters in the PPDS module. For the ADMS module, there is little change in model performance when $\beta$ is between 0.2 and 0.8, while there is a decrease in model performance at $\beta$ = 0 or 1, suggesting that both residual and cosine similarity terms are beneficial for model training.

\begin{figure}
	\begin{subfigure}{0.495\linewidth}
            \includegraphics[width=1\linewidth]{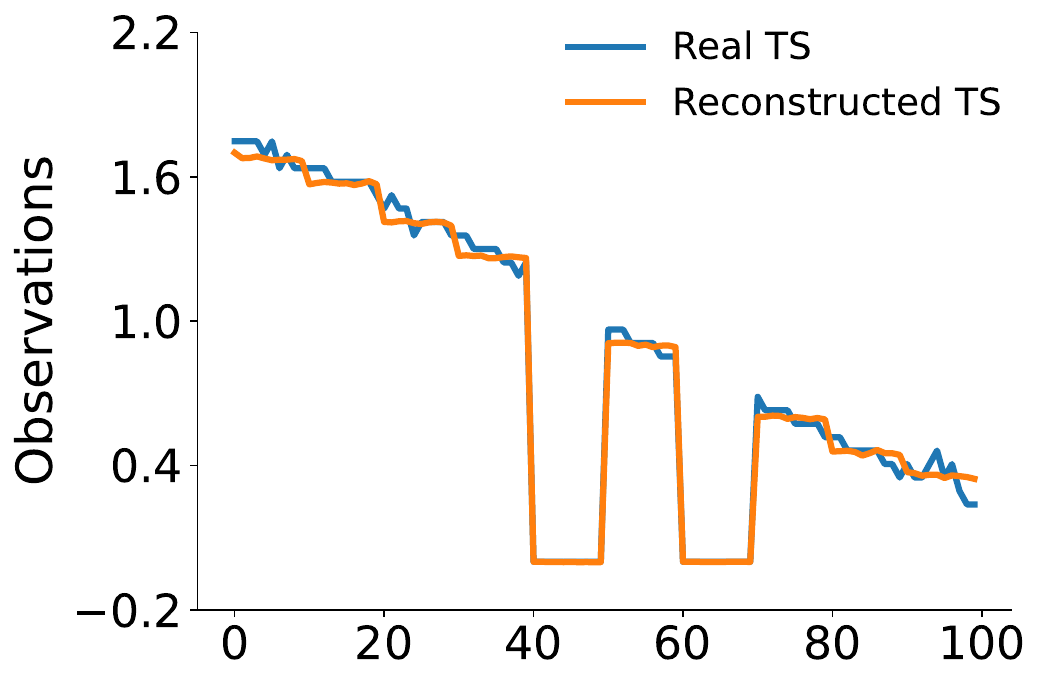}
            \captionsetup{font=scriptsize}
		\caption{A reconstruction example in training}
        \label{fig:train_recon}
	\end{subfigure}
	\begin{subfigure}{0.495\linewidth}
		\centering
		\includegraphics[width=1\linewidth]{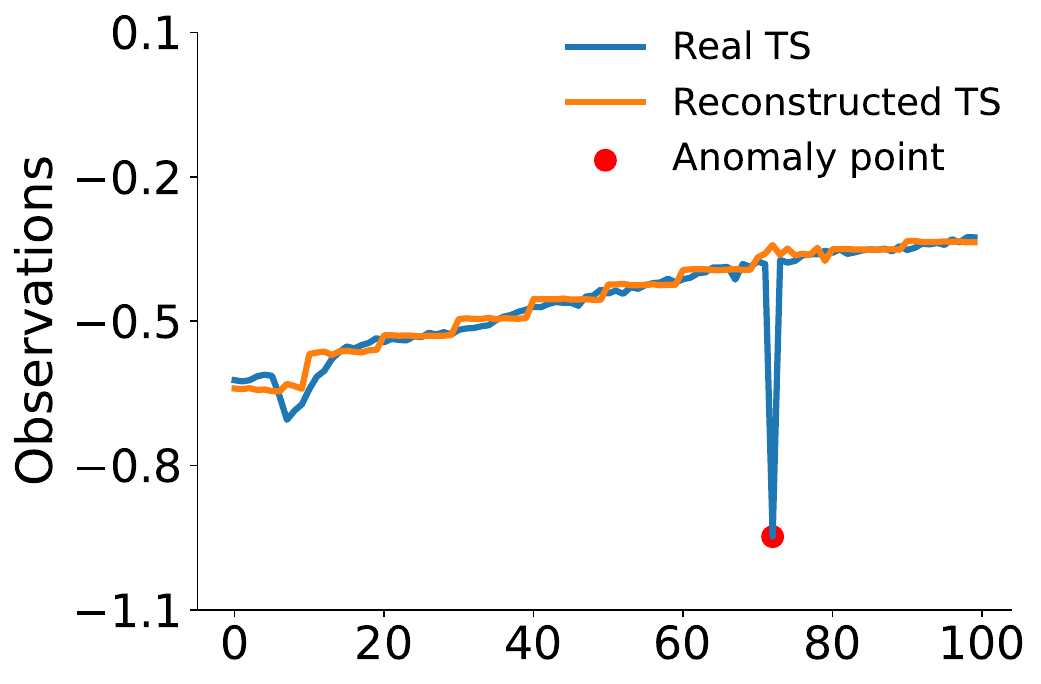}
              \captionsetup{font=scriptsize}
		\caption{A reconstruction example in testing}
        \label{fig:test_recon1}
	\end{subfigure}

        \captionsetup{font=small}
	\caption{Examples of time series reconstruction and anomaly detection within the client from SMD dataset. }
	\label{fig:appendix_case}
\end{figure}
\subsubsection{Case Study.}
We visualized two samples from the training and testing process and their reconstructed time series, respectively. Figure~\ref{fig:appendix_case} shows examples of series reconstruction during training and anomaly detection on the test data within the client. During training, the reconstructed curve almost matches the original time series. 
In testing, the estimated values at normal points closely approximate the true values, while at anomalous points, the estimates align more closely with reasonable values unaffected by anomalies. Thus the anomalies in the series are successfully identified by assessing the disparity between estimated and actual values.

\begin{table}[!t]
    \centering
    \caption{Comparison of Resources Resumption.}
    \label{tab:resources}
    \begin{tabular}{lcccc}
        \toprule
        & {\makecell{Comp Cost\\ (GFLOPS)}} & \makecell{Training \\ Time (s)} & \makecell{Memory\\ (Mb)} \\
        \midrule
    TimesNet$_{fl}$ & 319.22 & 131.63 & 427.60 \\
    FPT$_{fl}$ & 0.22 & 114.67 & 5594.50 & \\
    AT$_{fl}$ & 15.43 & 95.61 & 7875.00 & \\
    PeFAD$_{fl}$ & 0.43 & 57.22 & 2569.80 & \\
        \bottomrule
    \end{tabular}
\end{table}
\subsubsection{Resource Consumption}
We conduct experiments to compare the clients' resource consumption with the best performing baselines. The results on SMD dataset are shown in Table~\ref{tab:resources}. The results show that PeFAD has low training and computation costs, while other baselines fail to obtain a good balance between them.

\begin{table}[!t]
    \centering
    \caption{Continues Learning.}
    \label{tab:CL}
    \begin{tabular}{lcccc}
        \toprule
        & {\makecell{M1->MSL}} & \makecell{M1->PSM} & \makecell{M2->PSM} & \makecell{M2->MSL} \\
        \midrule
    AUC & 92.6 & 97.8 & 98.0 & 91.3 \\
    F1-Score & 78.9 & 97.3 & 97.4 & 77.4 \\
        \bottomrule
    \end{tabular}
\end{table}

\subsubsection{Continuous Learning}
We add a continuous learning (CL) experiment to assess PeFAD's performance on dynamic time series. The model is first trained on MSL dataset to obtain model M1 and then fine-tuned on PSM to get M2. We test whether M2 effectively learns new data (M2→PSM) while retaining old knowledge (M1→MSL).
The result is shown in Table~\ref{tab:CL}. It can be observed that PeFAD works well in CL scenarios due to the powerful generalization capabilities of PLM. Further, the fine-tuned PeFAD model performs well on PSM without forgetting knowledge of MSL, addressing catastrophic forgetting.

\end{document}